\documentclass[]{bytedance_seed}

\usepackage{minitoc}
\usepackage{url}
\usepackage{amssymb}
\usepackage[utf8]{inputenc}
\usepackage{microtype}
\usepackage{booktabs}
\usepackage{algorithm}
\usepackage{algpseudocode}
\usepackage{pifont} 
\usepackage{multirow}
\usepackage{makecell}
\usepackage{paralist}
\usepackage{xspace}
\usepackage{color}
\usepackage{xcolor}
\usepackage{colortbl}
\usepackage{adjustbox}
\usepackage{hyperref} 
\usepackage[edges]{forest}
\usepackage{tikz} 
\usepackage{caption}
\usepackage{amsfonts}
\usepackage[toc,page,header]{appendix}
\usepackage[utf8]{inputenc}
\usepackage{bbm}
\usepackage{enumitem}
\usepackage{amsthm}
\usepackage[dvipsnames]{xcolor}
\usepackage{colortbl}
\usepackage{array}

\definecolor{seedc}{RGB}{7, 92, 173}

\newcommand{\name}[1]{GR-3}
\newcommand{\Name}[1]{F-ACIL}

\newcommand{\hardware}[1]{ByteMini}

\renewcommand{\paragraph}[1]{\vspace{0.1em}\noindent\textbf{#1}}

\title{Towards Generalizable Robotic Data Flywheel: High-Dimensional Factorization and Composition}

\author[*,\dagger]{Yuyang Xiao}
\author[*]{Yifei Zhou}
\author{Haoran Wang}
\author{Wenxuan Ou}
\author[\dagger]{Yuxiao Liu}

\affiliation[]{ByteDance Seed}

\contribution[*]{Equal contribution}
\contribution[\dagger]{Corresponding authors}

\abstract{
The lack of sufficiently diverse data, coupled with limited data efficiency, remains a major bottleneck for generalist robotic models, yet systematic strategies for collecting and curating such data are not fully explored.
One major challenge in achieving generalization lies in the inherently high-dimensional nature of robotic data. Task diversity arises from implicit factors that are sparsely distributed across multiple dimensions and are difficult to define explicitly.
To address this challenge, we introduce \textbf{\Name{}}, a heuristic \textbf{F}actor-\textbf{A}ware \textbf{C}ompositional \textbf{I}terative \textbf{L}earning framework for data factorization and compositional generalization. 
\Name{} decomposes the data distribution into structured factor spaces such as object, action, and environment.
Based on the factorized formulation, we develop a factor-wise data collection and an iterative training paradigm that promotes compositional generalization over the high-dimensional factor space, leading to more effective utilization of real-world robotic demonstrations.
With extensive real-world experiments, we show that \Name{} can achieve more than 45\% performance gains with 5–10× fewer demonstrations comparing to that of which without the strategy.
The results suggest that structured factorization offers a practical pathway toward efficient compositional generalization in real-world robotic learning.
We believe \Name{} can inspire more systematic research on building generalizable robotic data flywheel strategies.
More demonstrations can be found at our website: \url{https://f-acil.github.io/}

}

\date{\today}
\correspondence{%
\email{xiaoyuyang.derek@bytedance.com}, \email{liuyuxiao.876@bytedance.com}%
}

\begin{document}
\maketitle

\section{Introduction}
\label{sect:intro}

\begin{figure}[t]
    \centering
    \begin{subfigure}[b]{0.47\textwidth}
        \centering
        \includegraphics[width=\textwidth]{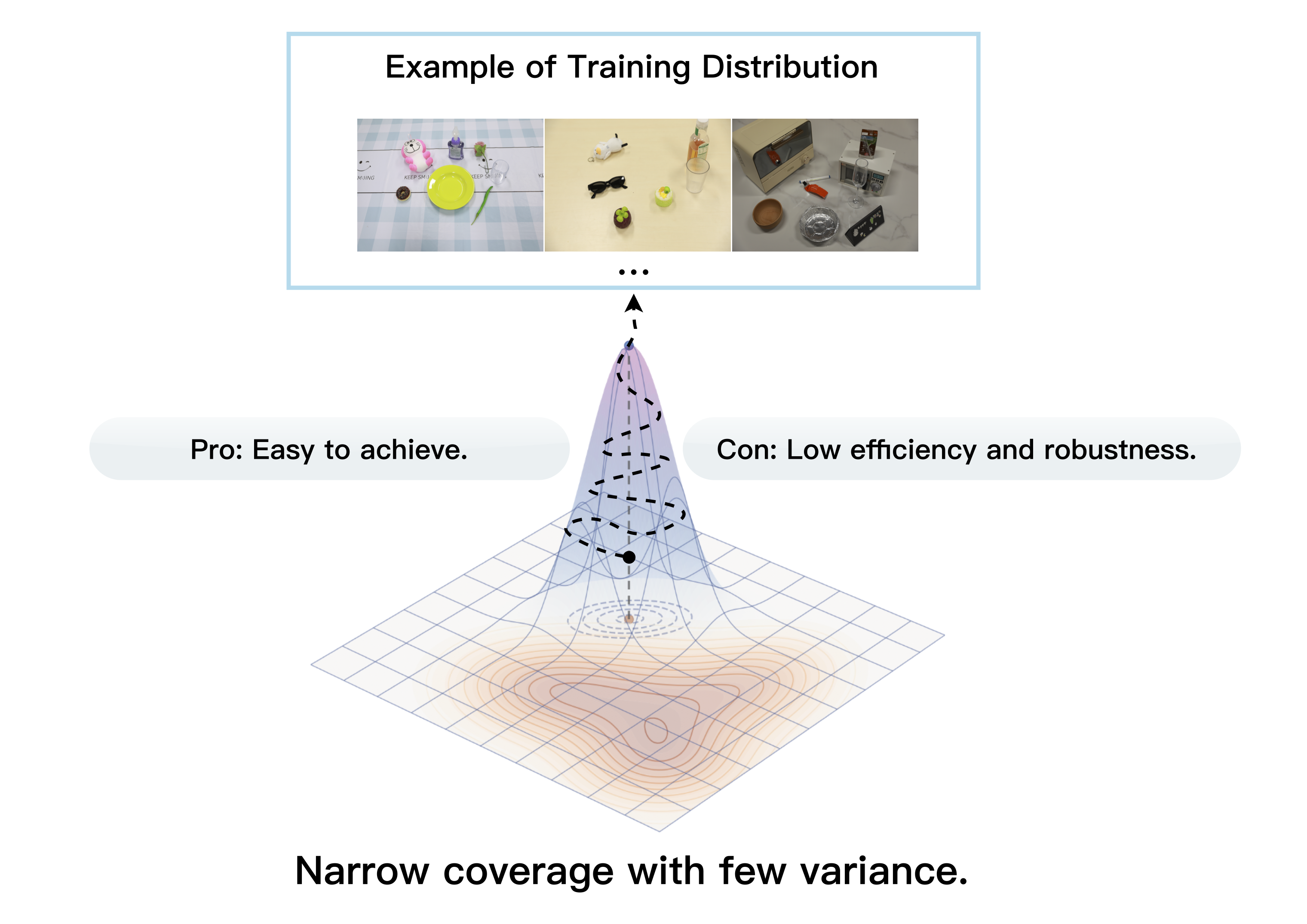}
        \caption{Gaussian-like}
        \label{fig:dist-biased}
    \end{subfigure}
    \hfill
    \begin{subfigure}[b]{0.47\textwidth}
        \centering
        \includegraphics[width=\textwidth]{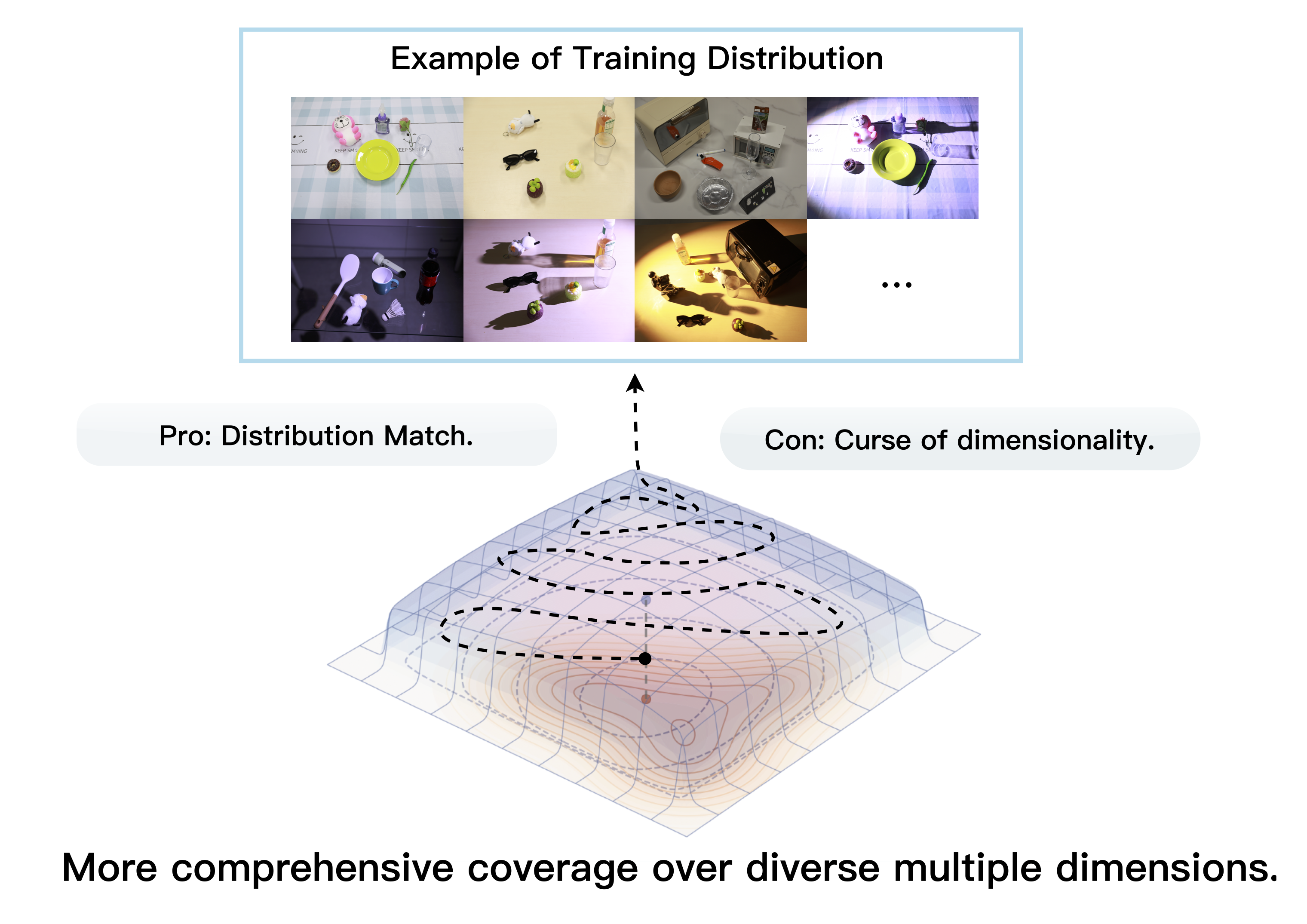}
        \caption{Quasi-uniform}
        \label{fig:dist-uniform}
    \end{subfigure}
    \\[1.2em]
    \begin{subfigure}[b]{0.19\textwidth}
        \centering
        \phantom{}
    \end{subfigure}
    \hfill
    \begin{subfigure}[b]{0.54\textwidth}
        \centering
        \includegraphics[width=\textwidth]{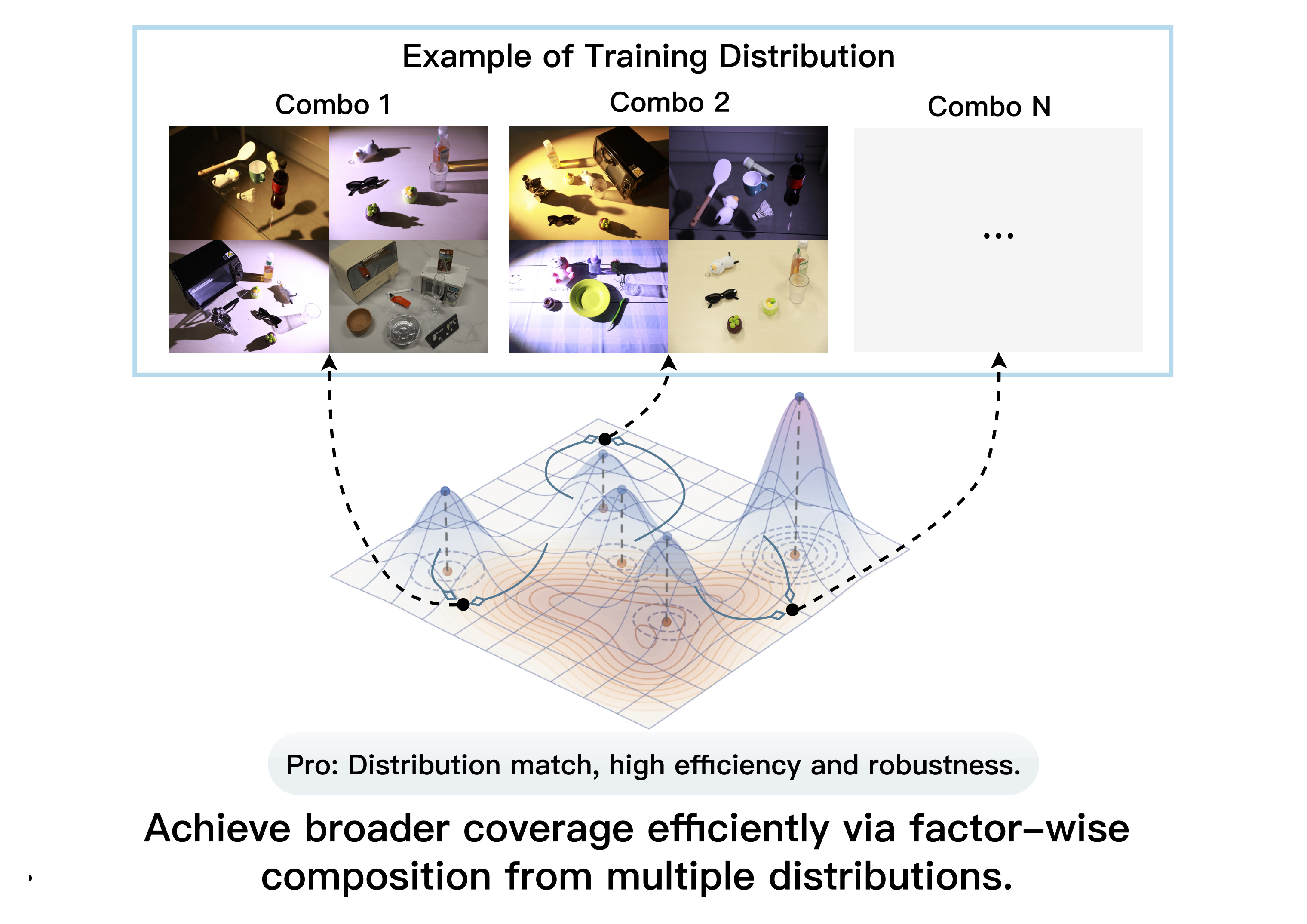}
        \caption{\Name{}: Sparse Gaussian Mixture}
        \label{fig:dist-mixture}
    \end{subfigure}
    \hfill
    \begin{subfigure}[b]{0.25\textwidth}
        \centering
        \includegraphics[width=\textwidth]{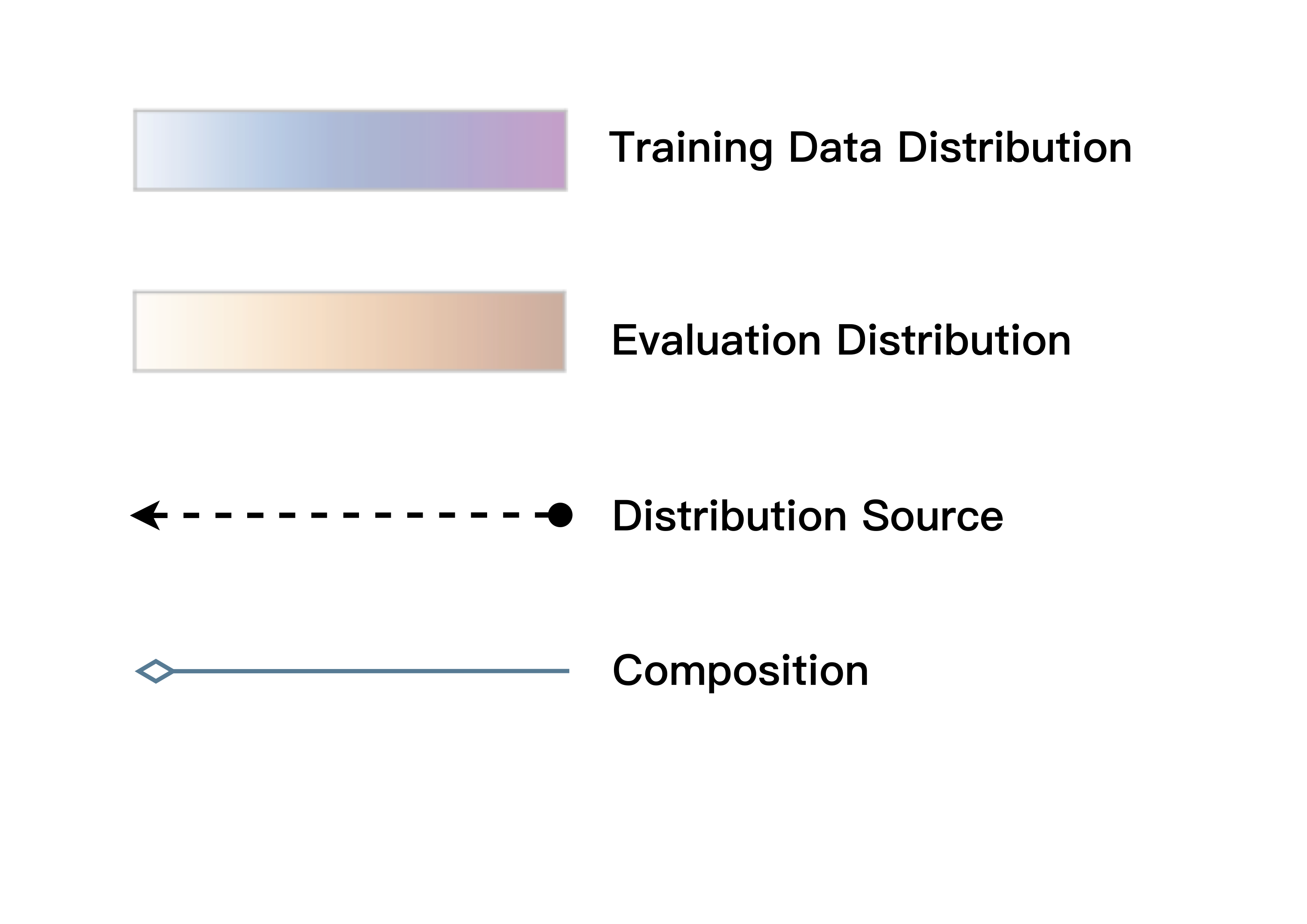}
    \end{subfigure}
    \caption{
    \textbf{Illustration of data distribution with specific properties.}
    (1)The 3D surfaces (blue-purple) represent different training data distributions, while the contour maps(warm amber) show the shared evaluation distribution.
    (2)We compare three data distribution: (a) a narrow Gaussian-like distribution with limited coverage, (b) a quasi-uniform distribution with full coverage but low efficiency, and (c) F-ACIL with multiple gaussian modes, which achieves efficient, broad coverage via factor-wise composition.
    }
    \label{fig:distributions}
\end{figure}

\begin{figure}[!t]
    \centering
    \vspace{-0.2cm}
    \includegraphics[width=\linewidth]{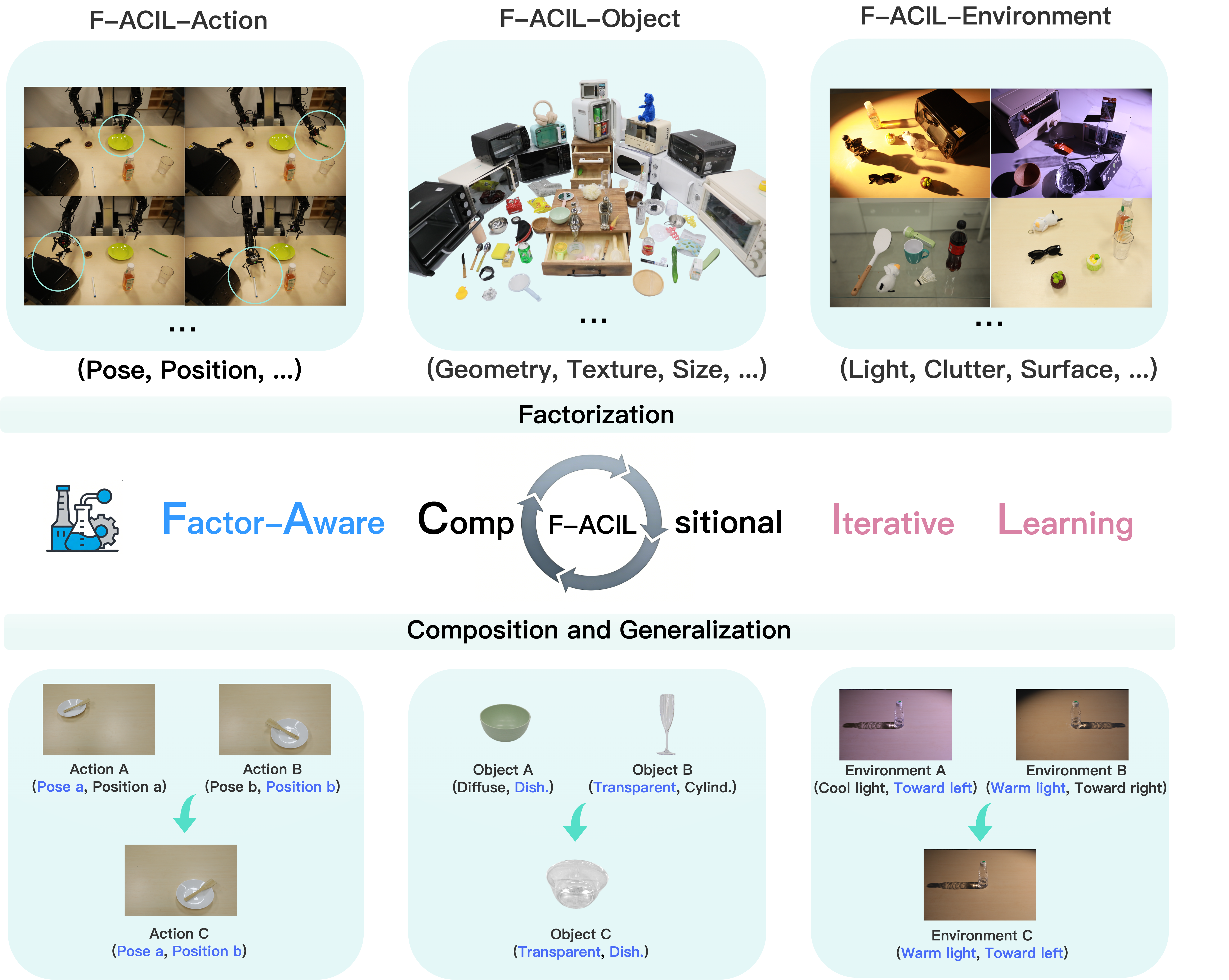}
    \caption{
        \textbf{Overview.}
        (1)\Name{} explicitly decomposes robotic manipulation into three factors: Object, Action, and Environment. This factorized representation is introduced to systematically expand the distribution of training data in real-world robotic learning.
        (2)\Name{} with factorized representation allows principled compositional generalization to out-of-distribution (OOD) instances.
    }
    \label{fig:teaser}
\end{figure}

\begin{figure}[!t]
    \centering
    \includegraphics[width=\linewidth]{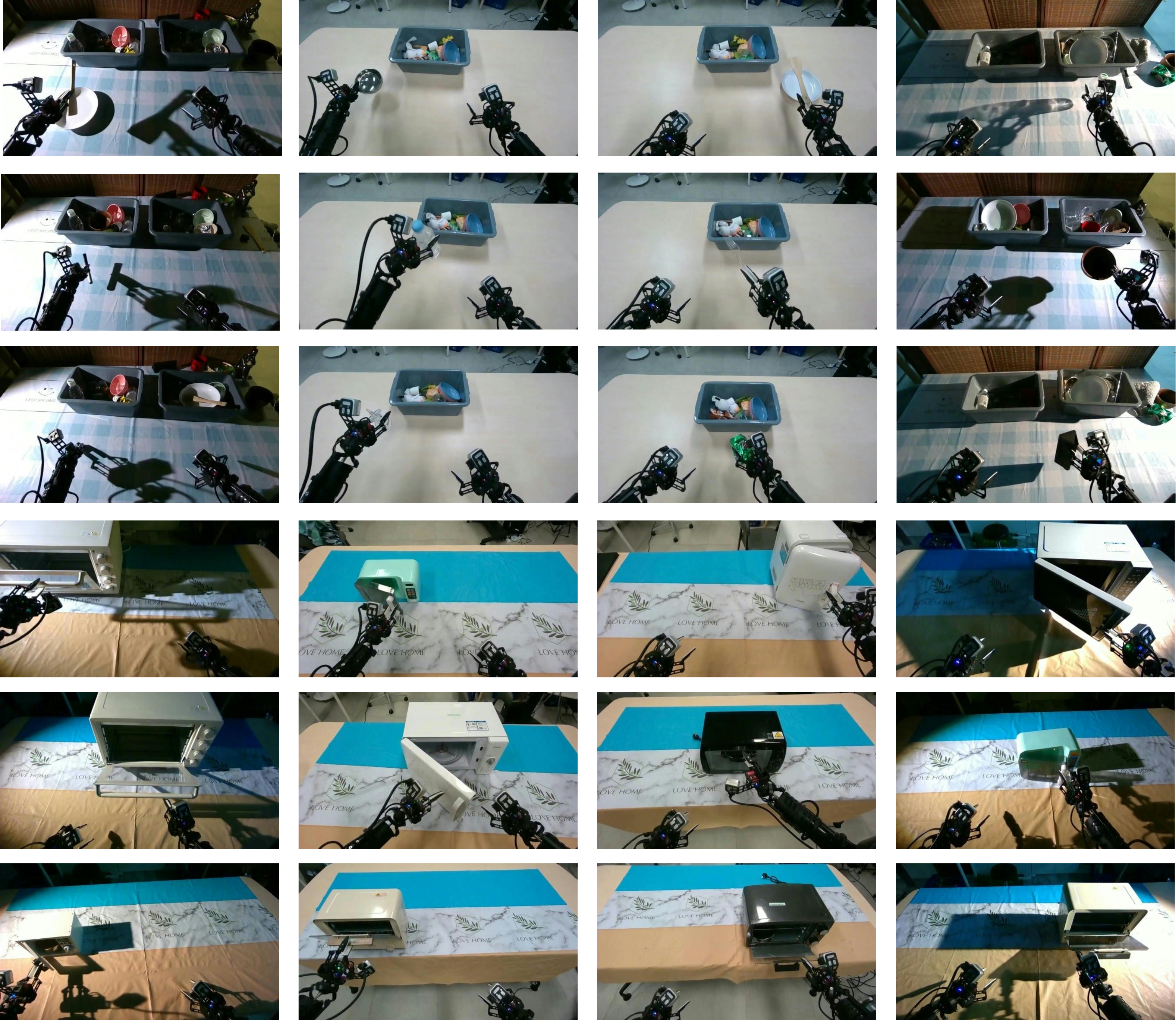}
    \caption{\textbf{Capabilities}. In both Pick-and-Place and Open-and-Close, all cases exhibit robust compositional generalization across out-of-domain objects, actions, and environments.}
    \label{fig:first_view}
\end{figure}

Data is the fuel of intelligence.
For vision-language-action models(VLAs) in robotic manipulation, performance is fundamentally constrained by the coverage and structure of training data.~\cite{cheang2025gr3technicalreport, fei2025liberoplus}. 
In practice, real-world demonstrations are collected with limited variations~\cite{huang2025adc, liu2023liberobenchmarkingknowledgetransfer, fei2025liberoplus}, which results in a long-tailed data distribution.
For instance, not only objects are typically arranged within a small set of fixed layouts, the action sequences in long-horizon tasks are also constrained by the collectors’ habitual patterns formed through repeated daily work, and so forth. 
Furthermore, real-world demonstrations are frequently gathered under convenient settings, ultimately causing distribution shift from training to real-world distribution.
As a result, the overall data coverage typically forms a Gaussian-like distribution, as illustrated in Fig.~\ref{fig:dist-biased}.
Such imbalanced distributions limit the robustness of learning-based models, hindering compositional generalization and degrading performance.

To overcome the coverage bias introduced by Gaussian-like data distributions, recent efforts build more diversified datasets with a Quasi-uniform distribution.(see Fig.~\ref{fig:dist-uniform}). 
Importantly, A fundamental challenge is to identify the axes of variation in the data distribution~\cite{gao2026taxonomy, fei2025liberoplus}, i.e., the specific factors along which diversification is required to ensure comprehensive coverage.
For instance, the works of ~\cite{zhou2025factorhd, fei2025liberoplus, liu2023liberobenchmarkingknowledgetransfer} introduce structured variations through underlying factors (e.g., object geometry, appearance, illumination, and background context), promoting balanced coverage of robotic data distributions and improving generalization.
However, naive expansion toward a quasi-uniform distribution across multiple axes in high-dimensional factor spaces often leads to the curse of dimensionality. 
While this approach increases data diversity, it suffers from low data efficiency and ultimately restricts effective generalization for robotic learning.

Hence, the distribution required for effective generalization is neither Gaussian-like nor Quasi-uniform.
In this work, we introduce \Name{} for factor-aware compositional iterative learning. As illustrated in Fig.~\ref{fig:dist-mixture}, we model the data distribution as the sparse mixture of Gaussian-like modes, where each mode captures coherent variations. By leveraging structured compositions across multiple modes in high-dimensional factor spaces, we believe that model can generalize beyond individual distribution and significantly reduce the need for exhaustive uniform coverage.
In general, Fig.~\ref{fig:teaser} provides an overview of \Name{}, which consists of structured axes for factorization and mechanisms for compositional generalization.
\Name{} maps the robotic states onto these structured axes: \Name{}-Object, \Name{}-Action, and \Name{}-Environment.  Regarding the process of compositional generalization, \Name{} reduces the complexity of exploring high-dimensional spaces and efficiently extends the generalization beyond the observed data distribution, and thus alleviating the need for exhaustive uniform coverage and improving data efficiency.

Overall, we use \Name{} to achieve compositional generalization across high-dimensional spaces and demonstrate its effectiveness on two representative manipulation skill families: \textbf{Pick-and-Place} and \textbf{Open-and-Close} (Fig.~\ref{fig:first_view}).
We empirically compare \Name{} with strategy that do not incorporate factor-aware iterative learning. The results show that \Name{} achieves over 45\% performance improvement while requiring 5–10× fewer demonstrations, highlighting its substantial gains in data efficiency and generalization. Furthermore, we find that effective compositional generalization relies on mixtures of factors with appropriate ratios, which are progressively refined through an iterative data flywheel.

\section{Related Work}
\label{sec:related_work}


\subsection{Factor-aware Robotic Evaluation}

\paragraph{Robotic Manipulation Benchmark.} Recent works on building robotic benchmark~\cite{james2020rlbench, guruprasad2024benchmarkingvisionlanguage, sedlacek2025realmrealtosimvalidatedbenchmark} emphasize task diversity and semantic generalization. In these settings, visual backgrounds, illumination, and camera configurations~\cite{liu2023liberobenchmarkingknowledgetransfer, zhou2025factorhd, fei2025liberoplus} are typically fixed or only slightly perturbed, providing a controlled but limited evaluation of policy robustness. Thus, several studies ~\cite{fei2025liberoplus, zhang2025experiencesbenchmarkingvisionlanguageactionmodels, zhou2025factorhd} strive to refine generalization metrics and evaluation protocols, introducing more systematic perturbations and standardized benchmarks for comparative analysis. 
While these benchmarks enhanced reproducibility and facilitate fair comparisons, the limited robotic actions and environments they produced constrain~\cite{zhang2025vlaarena, gao2026taxonomy} the systematic analysis of factor-wise policy adaptation. 
%
%
Therefore, we introduce \Name{}, which helps us perform systematic analysis by organizing robotic data into structured object, action, and environment factors.

\paragraph{VLAs With Real-World Evaluation.}  
Vision‑Language‑Action models(VLAs) aim to unify visual perception, language understanding, and action generation~\cite{ebert2021bridge, wen2024tinyvla, li2023vision} within a single multi-model framework, enabling generalist robot policies that follow natural instructions and interact with diverse environments. Early works such as RT‑2~\cite{brohan2023rt} study how internet‑scale vision‑language pretraining can be fine-tuned with robotic trajectory data to transfer semantic understanding to action spaces. Approaches that fuse vision–language models (VLMs) with action experts~\cite{ye2025vlar1, fang2025intentionexecutionprobinggeneralization} further improve reasoning and control by integrating reinforcement learning objectives and structured reasoning modules.
Despite these advances, VLA evaluations~\cite{lu2025vlarl, fei2025liberoplus, liu2023liberobenchmarkingknowledgetransfer, kang2023efficient} often largely focus on aggregate task success or semantic intent, without systematically disentangling how individual factor(e.g., object properties, action element, and environmental conditions) affects generalization performance.
In contrast, \Name{} directly evaluates how VLAs adapt to compositional variations, enabling a more precise understanding of how policies manipulate objects across diverse environments and various poses.
Moreover, \Name{} helps to provide insights on how different factors influence generalization robustness.

\subsection{Compositional Iterative Learning}

\paragraph{Compositional Generalization.}
Compositional generalization enables models to systematically composite learned primitives for novel tasks\cite{li2025theoreticalanalysiscompositionalgeneralization,ito2022compositionalgeneralizationabstractrepresentations, lecuyer2022tailoredvertexorderingfaster}.
Works focus on VLMs~\cite{chen2023pali, qu2025spatialvla, li2024context} reveal persistent gaps in compositional reasoning under out-of-distribution conditions, highlighting the need for explicit cross-modal grounding mechanisms.
Diffusion-based visuomotor policies~\cite{chi2024diffusionpolicy} have been proposed to generate multi-model action distributions, achieving strong performance across a wide range of manipulation tasks. Wang et al.~\cite{wang2024poco} employs a latent policy space where repertoire elements are interpolated, synthesizing behaviors compositionally from varied sources.
Furthermore, ~\cite{li2025unveil, mendez2022composuite} state the importance of cross-task and cross-modal reasoning capabilities, pointing out that existing VLMs still face significant weakness in compositional generalization, and emphasizing the importance of vision-language alignment and accurate visual grounding. 
Similarly, Driess et al.~\cite{driess2023Learning} propose compositional object-centric dynamic models that represent scenes as a set of independent object latents, generalizing to environments with varying numbers of objects.
Investigations into the joint development of language and sensorimotor representations~\cite{vijayaraghavan2024development, wang2023program, xu2025setup} have shown that increasing variation in task components enhances generalization to unseen verb–noun compositions in robotic systems.
Moreover, ~\cite{wu2022zeroshot, zhou2022policycompose, li2024context} suggests that task representations learned via internet data facilitate zero-shot adaptation by leveraging policy conditioning on novel combinations of disentangled factors.
We introduce \Name{}, a factor-wise robotic learning framework that exploits compositional structures inherent in observed distribution to enable efficient generalization. Instead of exhaustive enumeration over all possible factors, \Name{} iteratively expands along selected dimensions, leveraging relationships among different spaces to extrapolate to unseen compositions with fewer demonstrations, shifting generalization from passive coverage-based scaling to active dimension-aware expansion.

\paragraph{Data-efficient Robotic Learning.}
Improving data efficiency has become a central challenge in robotic learning, as collecting real-world robotic data is expensive and time-consuming.
Focusing on learning better visual and semantic representations, Gao et al.~\cite{gao2024efficient} demonstrate an efficient path of data acquisition, with variation along individual factors can yield strong compositional generalization if policies are able to exploit factor structure in the available data. The works of ~\cite{wu2022zeroshot, james2020rlbench, wang2025skil} suggest that task representations learned from semantic keypoints also efficiently facilitate zero-shot adaptation. In addition, Huang et al.~\cite{huang2025adc} substantially improves compositional generalization and robustness while requiring significantly less training data by compressing diverse task variations and failure-recovery behaviors into fewer episodes.
Several studies have demonstrated that structured inductive biases and object-centric decompositions significantly enhance learning efficiency. For instance, Zhang et al.~\cite{pmlr-v270-zhang25h} show that incorporating action locality into representation learning dramatically improves success rates on imitation tasks using only a handful of demonstrations. Similarly, Zhao et al.~\cite{zhao2025generalizable} state hierarchical object-centric skill decomposition enables strong generalization from minimal demonstrations by abstracting transferable skill primitives. Furthermore, hierarchical off-policy RL methods~\cite{nachum2018data, lu2025vlarl, hansen2021generalizationrlsoft} also exploit task decompositions to reduce required interactions. 
Beyond algorithmic improvements, recent works has also investigated data scaling and dataset diversity in imitation learning. Hu et al.\cite{hu2025datascalinglaws} show that the diversity of environments and objects plays a more important role than the absolute number of demonstrations in achieving robust generalization.
In addition, efforts in multitask and dexterous manipulation emphasize to begin improving data efficiency and mitigating problem of asymptotic optimization\cite{li2025spatialforce, cheang2025gr3technicalreport, ren2024learning}. 
However, these approaches rarely exploit the latent correlations within the dataset, leaving the potential for compositional generalization under limited data largely unexplored.
In this work, we propose \Name{}, which explicitly mines the correlation among multiple factors to enable compositional generalization in high-dimensional spaces and achieves substantially higher data efficiency than prior approaches. 
%
%
Importantly, \Name{} not only reduces the number of required demonstrations, but also preserves previously acquired reliable generalization capabilities.

\section{F-ACIL}
\label{sec:facil}
In this section, we formally outline the construction of F-ACIL. We begin by formulating the factorized state representation, and subsequently establish a compositional generalization framework.

\subsection{Factorized State Representation}
\label{sec:factot_system}

\paragraph{Factor System.}
Given an end-to-end vision--language--action policy:
\begin{equation}
\pi_{\theta} : \mathcal{S} \rightarrow \mathcal{A},
\end{equation}
its generalization capability is fundamentally governed by the coverage of the state space $\mathcal{S}$ that induced by the training distribution. However, $\mathcal{S}$ is typically high-dimensional and structurally entangled, making direct analysis impractical. To obtain a tractable and meaningful representation, we adopt a factorized perspective based on the physical structure of the real world. 

Motivated by the idea of structural decomposition, we approximate the state space as the cartesian product of three principal factor dimensions:
\begin{equation}
\mathcal{S} \approx \mathcal{O} \times \mathcal{A} \times \mathcal{E},
\end{equation}
where $\mathcal{O}$ denotes object-related factors, $\mathcal{A}$ captures action-related factors encoding spatial relationships, and $\mathcal{E}$ represents environmental factors.

The factor system provides a physically grounded abstraction of $\mathcal{S}$ that enables systematic analysis of coverage and generalization in large-scale VLA training. Model generalization is determined by the coverage of these factor spaces through factor-wise decomposition. To build intuition and make these ideas concrete, we now examine the definition of each factor space.

\paragraph{\Name{}-Object.}
Objects exhibit diverse properties, making exhaustive factor modeling impractical. 
We therefore select a small set of primary sub-dimensions that captures the most significant variations in visual observations. 
Specifically, we factorize object-related variations into three components: 
\textbf{texture}, \textbf{geometry}, and \textbf{size}, 
as summarized in Tab.~\ref{tab:object_factors}. The object factors considered in this work can be formulated as:
\begin{equation}
o = (t, g, s) \in \mathcal{O},
\end{equation}
where $t$, $g$, and $s$ denote the texture, geometry, and size of the object, respectively.

\begin{table}[t]
\centering
\caption{Object factors}
\label{tab:object_factors}
\begin{tabular}{llll}
\toprule
Factor & Dimension & Description & Example Objects \\
\midrule

\textbf{Texture}($T$)
& Transparent & Light transmission with refraction 
& Transparent water cup \\

& Specular & Mirror-like surface reflection 
& Aluminum can \\

& Diffuse & Rough surface with uniform reflection 
& Metal bowl, Foil tray \\

& Absorptive & Strong light absorption 
& Towel, Tape roll \\

\midrule

\textbf{Geometry}($G$)
& Cylindrical & Axis-symmetric geometry 
& Cup, Aluminum can \\

& Dish-like & Roughly concave.
& Tray, Plate \\

& Rod-like & Large aspect ratio & Knife, Fork \\

& Irregular & No specific shape & Towel, Tissue \\

\midrule

\textbf{Size}($S$)
& Small & Requires high grasp accuracy
& Tape roll \\

& Medium & Easy single-grasp objects & Water cup \\

& Large & Requires wider grasping volume  & Lay's Chips, Towel \\

\bottomrule
\end{tabular}
\end{table}
\paragraph{\Name{}-Action.}
Directly modeling action-related factors is challenging due to the high variability of robot demonstrations. For many manipulation tasks (e.g., "put $\star$ into $\star$"), the model is not required to follow a specific trajectory as long as the task is successfully completed. Therefore, we characterize action diversity through variations in the initial scene configuration.

Depending on our hardware configuration, we decide to analyze the 6-D spatial distribution of the demonstration. As shown in Tab.~\ref{tab:action_factors}, the action factor is defined as:
\begin{equation}
a = (x, y, z, \phi, \theta, \psi) \in \mathcal{A},
\end{equation}
where $(x,y,z)$ represents the object position and $(\phi, \theta, \psi)$ denotes the roll, pitch, and yaw angles.

\begin{table}[t]
\centering
\caption{Action factors}
\label{tab:action_factors}
\begin{tabular}{llll}
\toprule
Factor & Dimension & Description & Range  \\
\midrule
\textbf{Position}
& $x$ & Forward direction of the robot & workspace \\
& $y$ & Left direction of the robot & workspace \\
& $z$ & Up direction of the robot & workspace \\
\midrule
\textbf{Orientation}
& roll($r$) & Rotation around $x$-axis & $[-\pi, \pi)$ \\
& pitch($p$) & Rotation around $y$-axis & $[-\pi, \pi)$ \\
& yaw($y$) & Rotation around $z$-axis & $[-\pi, \pi)$ \\
\bottomrule
\end{tabular}
\end{table}

\paragraph{\Name{}-Environment.}
Environmental factors introduce substantial variations in visual observations and often affect the robustness of robotic manipulation. 
To model these variations in a structured manner, we divide environmental factors into two levels: \textbf{Macro} and \textbf{Micro}.

Macro-level factors describe global imaging conditions that influence the entire scene, such as shadow direction, lighting intensity, and color temperature. 
These variables primarily affect the overall features perceived by the model. Micro-level factors characterize properties of the local workspace. Comparing to macro factors, these elements mainly introduce localized visual changes around the manipulation region and may directly interfere with object perception.

Dividing environmental factors allows us to represent diversity with a small number of interpretable variables, while still capturing the major sources of variation. 
The resulting environmental factors are summarized in Tab.~\ref{tab:environment_factors}. Formally, we define the environmental factor as:
\begin{equation}
e = (i, t, d, m, c) \in \mathcal{E},
\end{equation}
where $i, t, d, m$ and $c$ denote the light intensity, color temperature, shadow direction, surface material and background clutter.

\begin{table}[!t]
\centering
\caption{Environment factors}
\begin{tabular}{llll}
\toprule
Factor & Dimension & Description & Examples\\
\midrule

\textbf{Macro}
& Light Intensity($I$) & Overall brightness of the scene & Normal Indoor(200-500 lux)\\

& Color Temperature($T$) & Warm or cool illumination & Warm Light(2700K - 3500K) \\

& Shadow Direction($D$) & Direction of the shadow & Left, Mid, Right  \\
\midrule
\textbf{Micro}
& Surface Material($M$) & The material of local scenes & Transparent, Absorptive\\

& Background Clutter($C$) & The level of clutter in the scene & Clean, Light Clutter, Heavy Clutter\\

\bottomrule
\end{tabular}
\label{tab:environment_factors}
\end{table}

\subsection{Factor-Wise Composition and Generalization}
\label{sec:compo_general}

In this part, we begin by formulating generalization in the single-factor case. 
We then extend the formulation to multi-factor scenarios. Finally, we present a data collection strategy guided by the factor-wise compositional generalization.

\subsubsection{Defining Factor-Wise Generalization}

\begin{figure}
    \centering
    \includegraphics[width=0.5\linewidth]{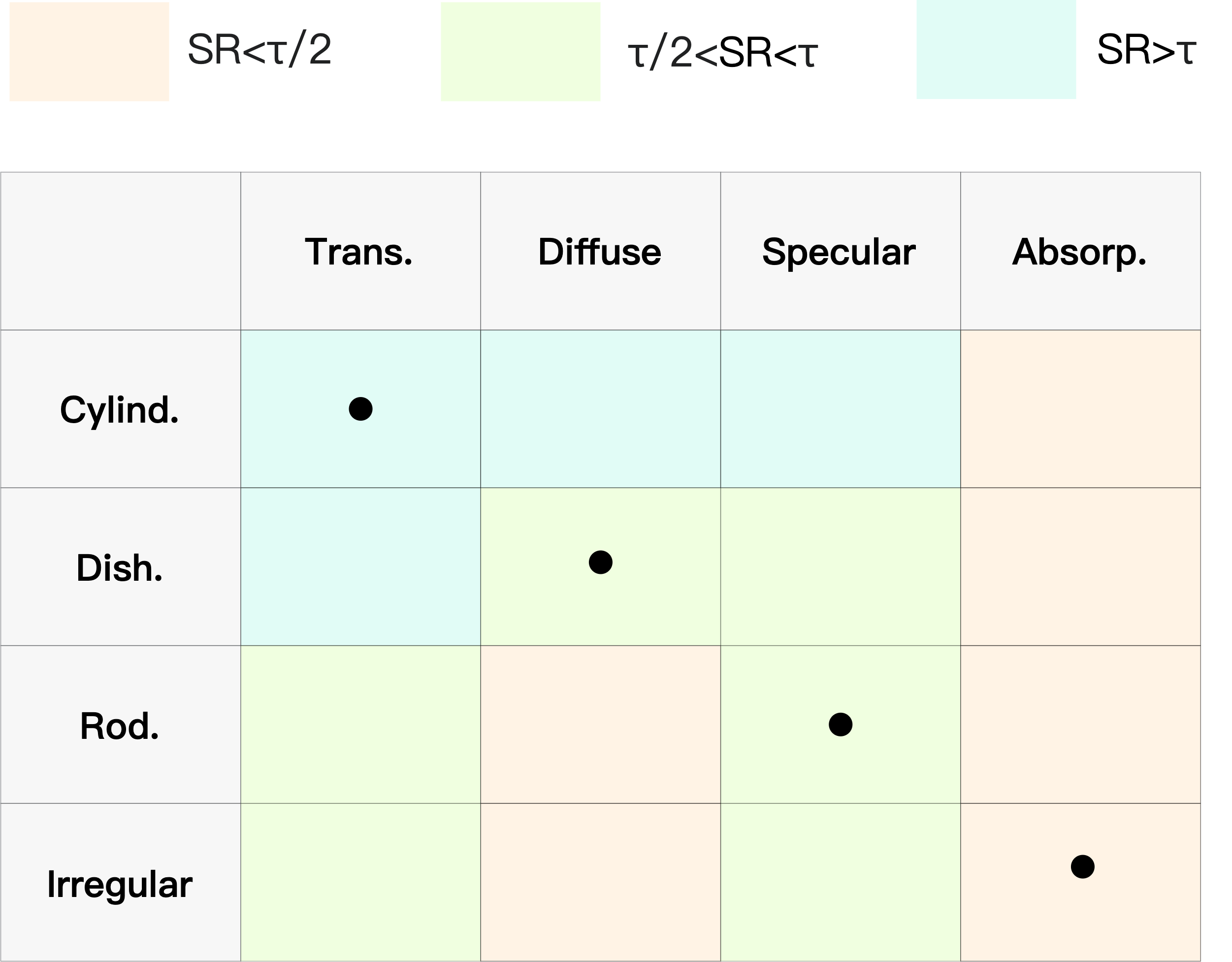}
    \caption{
        \textbf{Demonstrations collected for iteration 1 under $\mathcal{O}$ space}
        The initial demonstrations are sampled along the main diagonal of the factor space, with an equal number of samples per composition. The background color encodes the success rate(SR). Black dots mark the compositions included in the dataset, denoted as $D_1$.
    }
    \label{fig:iter1}
\end{figure}

\paragraph{Low-Dimensional Iteration.}
We formulate generalization along each factor dimension through a "group action"\footnote{This is an analogy to the mathematical definition of group actions. We define the term orbit in a similar spirit.} perspective. 
Taking the object factor space $\mathcal{O}$ as an example. First, we define a function $f$ that maps a dataset to the object factor space:

\begin{equation}
    f: D \to \mathcal{O}, \label{eq:DtoO}
\end{equation}

where $D$ denotes the dataset. $f(D)$ denotes the corresponding factor compositions under object factor. For example, the dataset mentioned in Fig.~\ref{fig:iter1}, denoted as $D_1$, 
is uniformly distributed over four factor compositions. Formally,
\begin{equation}
f(D_1) = [(\text{Transparent}, \text{Cylindrical}), (\text{Diffuse}, \text{Dish-like}), (\text{Specular}, \text{Rod-like}), (\text{Absorptive}, \text{Irregular})].
\end{equation}

Then we characterize the range of the factor space that a model can generalize to via an orbit closure induced by an empirical transformed set. Given a set of collected demonstrations $D$, $f(D) \subset \mathcal{O}$ denote the collected object factors. 
We define the orbit of $D$ under transformations $H_{\mathcal{O}}$ as
\begin{equation}
\mathrm{Orb}_{H_{\mathcal{O}}}(D)
:= 
\{\, g \cdot s \mid g \in H_{\mathcal{O}},\; s \in f(D) \,\},
\end{equation}
where $H_{\mathcal{O}}$ denotes the set of empirical transformations obtained from evaluation along $\mathcal{O}$.
Intuitively, $  \mathrm{Orb}_{H_{\mathcal{O}}}(D)  $ denotes the set of object factors to which a model trained on dataset $D$ can generalize. 

However, $f(D)$ does not necessarily lie within $  \mathrm{Orb}_{H_{\mathcal{O}}}(D)  $. As illustrated in Fig.~\ref{fig:iter1}, the $\mathrm{Orb}_{H_{\mathcal{O}}}(D_1)$ is the region where the success rate over $\tau$. Notably, the composition (Absorptive, Irregular) belongs to $f(D_1)$ but lies outside $\mathrm{Orb}_{H_{\mathcal{O}}}(D_1)$. This discrepancy reveals that certain factor compositions are inherently more difficult to learn than others. Thus, achieving reliable generalization requires a greater number of samples.

Our goal is therefore to identify a training dataset $D'$ so that
\begin{equation}
\mathrm{Orb}_{H_{\mathcal{O}}}(D') = \mathcal{O},
\end{equation}
while minimizing the cardinality\footnote{The number of unique elements in the set.} of its factor representation,
\begin{equation}
|f(D')| \ll |\mathcal{O}|.
\end{equation}

The definition implies that there are many factor compositions that are not in demonstrations, but the model can still generalize across the entire object factor space $\mathcal{O}$.

\paragraph{Multi-Dimensional Iteration.}
The definition of $\mathrm{Orb}_{H_{\mathcal{O}}}(D)$ can extends seamlessly to multi-factor spaces. For example, on the product space $\mathcal{O}\times\mathcal{A}$ (short for $\mathcal{O}\mathcal{A}$), we define the family of dataset that helps model to generalize across entire factor space as follows:
\begin{equation}
\{D \mid f(D) \subset \mathcal{O}\mathcal{A},\;
\mathrm{Orb}_{H_{\mathcal{O}\mathcal{A}}}(D)
=
\mathcal{O}\mathcal{A}
\}.
\end{equation}

Consequently, our goal is to identify a dataset from this family that achieves full generalization with a relatively small volume.

\subsubsection{Data Collection Strategy}
\label{sec:data_collection}

We adopt a hierarchical data collection strategy that includes
\begin{itemize}
    \item sequential factor exploration
    \item iterative subset search within each factor space
\end{itemize}
The strategy enables scalable exploration of the full factor space while mitigating combinatorial explosion.

\paragraph{Sequential Factor Expansion.}
As discussed in Sec.~\ref{sec:factot_system}, we prioritize the factor spaces according to their relative impact on generalization, following the order
\begin{equation}
\mathcal{O} \rightarrow \mathcal{A} \rightarrow \mathcal{E}. \label{eq:oae}
\end{equation}
According to Eq.~\ref{eq:oae}, the exploration of the factorized state representation is conducted in a sequential manner following the hierarchy, i.e., progressively expanding from $\mathcal{O}$ to $\mathcal{A}$ and finally to $\mathcal{E}$.

Starting from the object space $\mathcal{O}$, we first identify a dataset 
$D_\mathcal{O}$, with $ f(D_\mathcal{O}) \subset \mathcal{O}$ such that
\begin{equation}
\mathrm{Orb}_{H_{\mathcal{O}}}\bigl(D_\mathcal{O}\bigr) = \mathcal{O},
\end{equation}

Based on $f(D_\mathcal{O})$, we extend the search to the reduced product space $f(D_\mathcal{O})\mathcal{A}$, rather than directly considering the full cartesian product space $\mathcal{O}\mathcal{A}$. Searching the entire $\mathcal{O}\mathcal{A}$ space simultaneously would incur much higher computational cost and slow down the iterative process. Therefore, we approximate the object space using the compact representative set $f(D_\mathcal{O})$ in order to accelerate the exploration.

Let $D_\mathcal{OA}$ denote the empirically obtained minimal dataset satisfying
\begin{equation}
\mathrm{Orb}_{H_{f(D_\mathcal{O})\mathcal{A}}}\bigl(D_\mathcal{OA}\bigr) 
= f(D_\mathcal{O})\mathcal{A}.
\end{equation}
This implies that model trained on $D_\mathcal{OA}$ can generalize to the reduced space 
$f(D_\mathcal{O})\mathcal{A} \subset \mathcal{O}\mathcal{A}$. Although this condition does not guarantee the generalization to the full product 
$\mathcal{O}\mathcal{A}$, we further show (in Sec.~\ref{subsec:results}) that
\begin{equation}
\mathrm{Orb}_{H_{\mathcal{O}\mathcal{A}}}\bigl(D_\mathcal{OA}\bigr) 
= \mathcal{O}\mathcal{A}.
\end{equation}
Hence, model trained on $D_\mathcal{OA}$ generalizes to the entire object-action space $\mathcal{O}\mathcal{A}$. 
Similarly, in terms of coverage, exploring $\mathcal{O}\mathcal{A}$ is effectively equivalent to the sequential expansion from $f(D_\mathcal{O})$ to 
$f(D_\mathcal{O})\mathcal{A}$. 

Same procedure applies to the expansion from $f(D_\mathcal{OA})$ to $f(D_\mathcal{OA})\mathcal{E}$. Ultimately, we obtain an empirically minimal dataset $D_\mathcal{OAE}$ such that
\begin{equation}
\mathrm{Orb}_{H_{\mathcal{O}\mathcal{A}\mathcal{E}}}
\bigl(D_\mathcal{OAE}\bigr) 
= \mathcal{O}\mathcal{A}\mathcal{E},
\end{equation}
that is, a relatively compact representative set whose orbit under 
$H_{\mathcal{O}\mathcal{A}\mathcal{E}}$ covers the full factor space. 

Overall, the proposed sequential factor expansion achieves the same coverage as the direct exploration over the full cartesian product space, while requiring substantially less computational cost and data. 
Comprehensive empirical evidence is presented in the Sec.~\ref{sec:experiments}.

\begin{figure}[!t]
    \centering
    \includegraphics[width=0.8\linewidth]{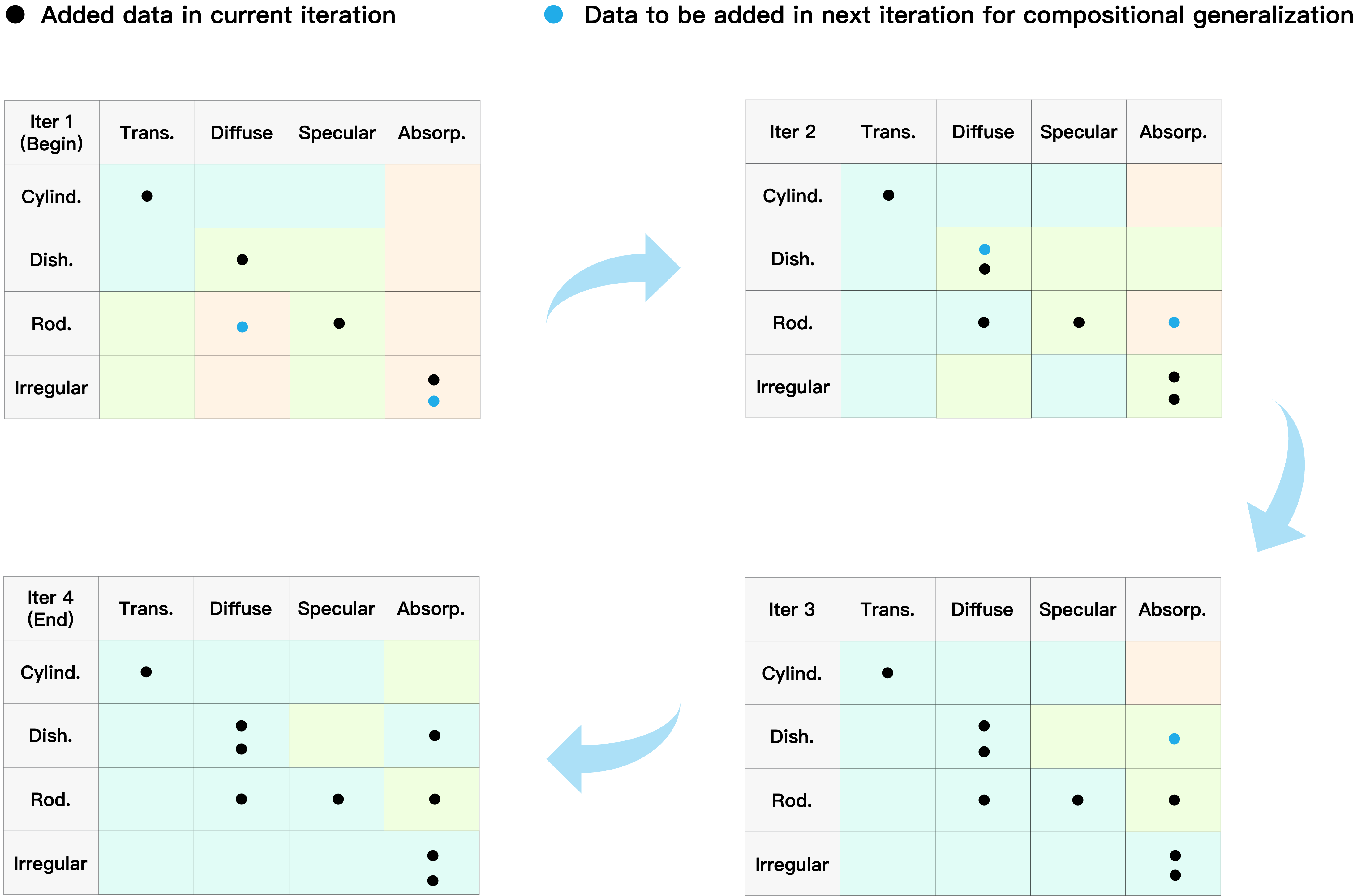}
    \caption{
        \textbf{Iterative Search Procedure in Object Space}
        In each iteration, different colors indicate the performance under specific factor compositions. The background color encoding follows the same scheme as in Fig.~\ref{fig:iter1}. Points highlighted in \textcolor{blue}{blue} mark factor compositions that are newly added into the demonstrations in the current iteration. The number of points reflects the distribution of ratio across different factor compositions.
    }
    \label{fig:iteration}
\end{figure}

\begin{algorithm}[t]
\caption{Iterative Factor-Based Data Expansion}
\label{alg:factor_object}
\begin{algorithmic}[1]
\Require Factor set $\mathcal{O}$, factor-based benchmark $\mathcal{B}_{\mathcal{O}}$, success threshold $\tau$, dataset unit size $n$

\State Construct $\mathcal{D}_{\text{diag}}$ by collecting data from hyper-diagonal factor configurations, with cardinality $n$ for each factor composition.
\State Train model $\pi_\theta$ on $\mathcal{D} = \mathcal{D_\text{diag}}$

\While{true}
    \State Evaluate $\pi_\theta$ on benchmark $\mathcal{B}_{\mathcal{O}}$
    \State Compute success rate for each factor $r(o), \forall o \in \mathcal{O}$, get success rate tensor $\mathcal{R}$
    \State Compute overall success rate 
    \begin{equation}
        r \leftarrow \frac{1}{|\mathcal{O}|} \sum_{o \in \mathcal{O}} r(o)
    \end{equation}
    \If{$r \geq \tau$}
        \State \textbf{break}
    \EndIf
    \State Update demonstrations $\mathcal{D}$ by Alg.~\ref{alg:curate_data}\Comment{Update dataset by factorization and composition}
    \State Retrain model $\pi_\theta$ on updated $\mathcal{D}$
\EndWhile

\State \Return $\pi_\theta$, $\mathcal{D}$
\end{algorithmic}
\end{algorithm}

\begin{algorithm}[t]
\caption{Collecting Data based on Factorization and Composition}
\label{alg:curate_data}
\begin{algorithmic}[1]
\Require Success tensor $\mathcal{R}$, factor set $\mathcal{O}$, selected data set $\mathcal{D}$, success threshold $\tau$, dataset unit size $n$ and function $f$ as defined in Eq.~\ref{eq:DtoO}

\State Compute aggregated tensor $\mathcal{S}$ from $\mathcal{R}$ by Eq.~\ref{eq:aggregated_tensor}
\State $\mathcal{M} \leftarrow \{ o \mid o \in \mathcal{O},r(o) > \tau \}$

\While{$\mathcal{M} \neq \mathcal{O}$} \Comment{Stop until every element in factor set is marked}
    \State $\mathbf{s} \leftarrow \arg\min_{o \in \mathcal{O} \setminus \mathcal{M}} \mathcal{S}$ \Comment{Identify the poorly performed factor across all dimensions}
    \State Mark $\mathbf{s}$ and update $\mathcal{M} \leftarrow \mathcal{M} \cup \{\mathbf{s}\}$
    
    \ForAll{$\mathbf{d} \in f(\mathcal{D})$}
        \State Mark all vertices of the hypercube spanned by $\{\mathbf{s}, \mathbf{d}\}$
        \State Update $\mathcal{M}$ accordingly
    \EndFor
    \State Collect dataset $\Delta D$ such that $f(\Delta D) = \{\mathbf{s}\}$ and $|\Delta D|=n$
    \State $\mathcal{D} \leftarrow \mathcal{D} \cup \Delta D$ 
\EndWhile

\State \Return $\mathcal{D}$
\end{algorithmic}
\end{algorithm}

\paragraph{Iterative Subset Search.}\label{para:iterative_search}
Within each factor space, we employ an iterative search procedure, as illustrated in Fig.~\ref{fig:iteration}, to identify a minimal dataset with sparse factor representations (e.g., $D_\mathcal{O}$, $D_\mathcal{OA}$ and $D_\mathcal{OAE}$ mentioned above), whose induced orbit achieves full coverage of the corresponding space.

Taking the object space as an example, our algorithm begins by constructing an initial training dataset (e.g., $D_1$ in Fig.~\ref{fig:iter1}) composed of samples with diagonal factor configurations. A model is then trained on this dataset.

Then, the model is evaluated on a factor-based benchmark specifically designed to measure performance across different object factors. From the evaluation, we obtain the individual success rate for each object factor and the overall success rate aggregated across all factors, resulting in an success rate tensor $\mathcal{R}$.

We define tensor $\mathcal{S}$ as follows: 
\begin{equation}
S_{\mathbf{i}}
=
\sum_{m=1}^{n}
\sum_{\mathbf{j} \,:\, j_m = i_m}
R_{\mathbf{j}}
-
(n-1) R_{\mathbf{i}},
\label{eq:aggregated_tensor}
\end{equation}
where $\mathbf{i}=(i_1,\dots,i_n)$, $\mathbf{j}=(j_1,\dots,j_n)$. 
We identify poorly performed factor compositions and augment the training data accordingly. Fig.~\ref{fig:iteration} shows how we adopt a compositional generalization paradigm to selectively choose samples from poorly performing factor space regions. 
The key idea is that by learning from a limited number of carefully chosen compositions, our model can capture transferable structural regularities, and therefore is able to generalize to unseen factor compositions. The model is then retrained on the expanded dataset, and the process is repeated until the overall success rate exceeds the threshold $\tau$. The complete procedure is summarized in Alg.~\ref{alg:factor_object} and Alg.~\ref{alg:curate_data}.

The iterative search procedure over the reduced product space $  f(D_\mathcal{O})\mathcal{A}  $ is conceptually analogous to that over the object space $  \mathcal{O}  $. 
In the object-space search, we obtain a dataset $  \mathcal{D}_\mathcal{O}  $ which ensures full generalization across $  \mathcal{O}  $. When we subsequently apply the same iterative procedure to $  f(D_\mathcal{O})\mathcal{A}  $, the mapping $  f(D_\mathcal{O})  $ remains fixed throughout the process. 
As the result, the search procedure is equivalent to optimizing the action-factor distribution while preserving the already-stabilized object-factor coverage induced by $  \mathcal{D}_\mathcal{O}  $.
%
%
Such approach allows us to systematically explore the optimal factor ratio in the action space $  \mathcal{A}  $ on top of a high-quality base distribution over objects. 

The iterative search over $\mathcal{OAE}$ is performed accordingly.

\subsubsection{Compositional Generalization and Factor Design} 
Our data collection strategy is fundamentally built upon the principle of compositional generalization, which serves as the primary mechanism for achieving superior data efficiency.

\begin{figure}
    \centering
    \begin{subfigure}[b]{0.49\textwidth}
        \centering
        \includegraphics[width=\textwidth]{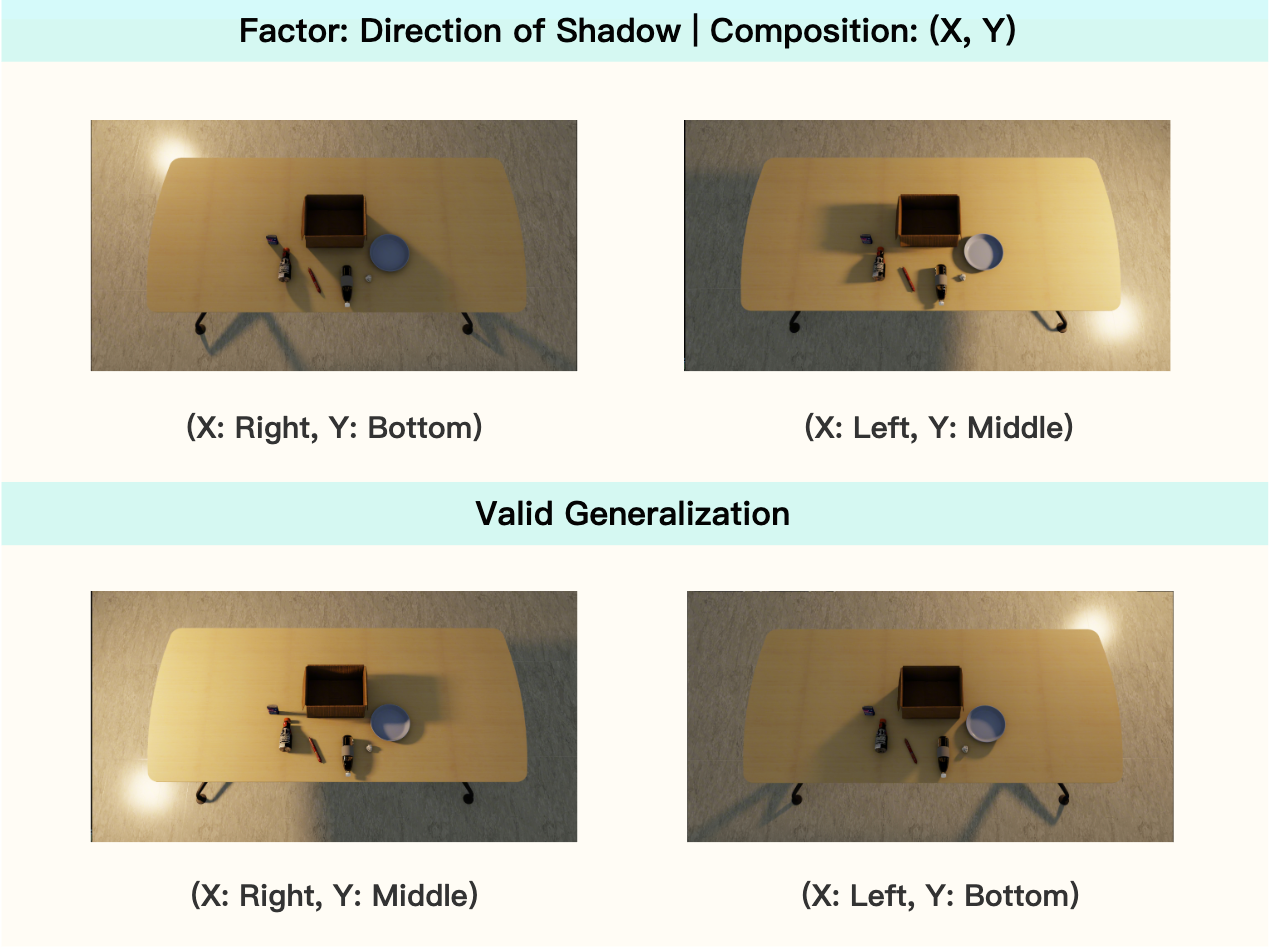}
        \caption{Valid Factor-wise Generalization}
        \label{fig:valid-gen}
    \end{subfigure}
    \hfill
    \begin{subfigure}[b]{0.49\textwidth}
        \centering
        \includegraphics[width=\textwidth]{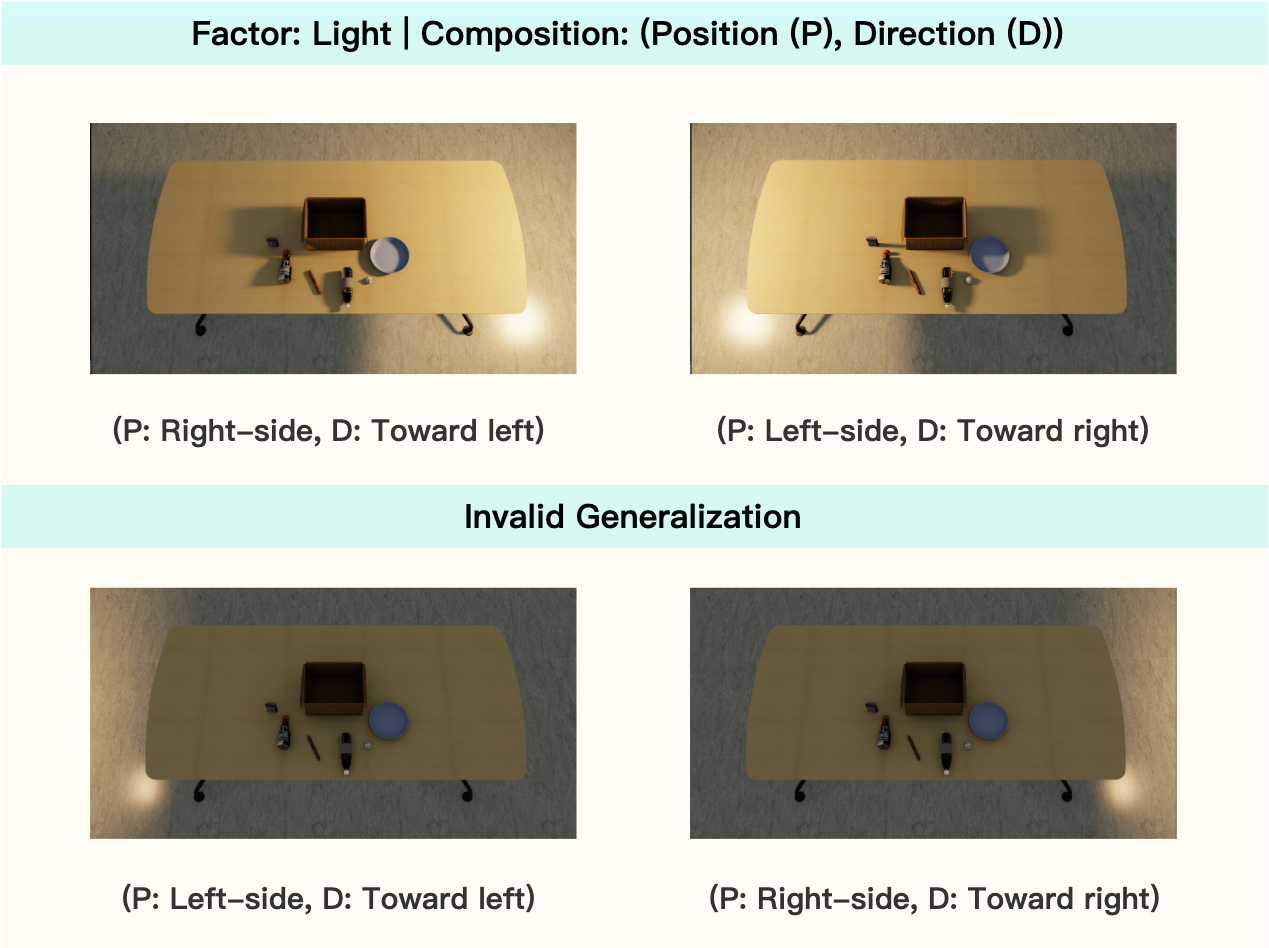}
        \caption{Invalid Factor-wise Generalization}
        \label{fig:invalid-gen}
    \end{subfigure}
    \vspace{1em}
    \caption{
        \textbf{Empirical Study on Factors Design} (a) Shadow direction constitutes a successful factor design, as strong compositional generalization is observed. (b) Light-source position and direction serve as a counterexample, where compositional generalization fails to hold.
    }
    \label{fig:q1}
\end{figure}

For example, the direction of shadow can be naturally treated as a well-defined factor. We represent this factor as a $3 \times 3$ matrix, where the rows $  \{\text{Top}, \text{Middle}, \text{Bottom}\}  $ discretize positions along the vertical direction of the tabletop (y-axis), and the columns $  \{\text{Left}, \text{Center}, \text{Right}\}  $ discretize positions along the horizontal direction of the tabletop (x-axis). As shown in Fig.~\ref{fig:valid-gen}, the observed factor combinations (X: Right, Y: Bottom) and (X: Left, Y: Middle) can successfully generalize, via compositional generalization, to the unseen combinations (X: Right, Y: Middle) and (X: Left, Y: Bottom).

However, compositional generalization between factors does not always hold. For example, the position and direction of the light source do not necessarily satisfy compositional generalization. We represent position and direction of the light as a $  2 \times 2  $ matrix with positions: $  \{ \text{Left-side}, \text{Right-side}\}  $ and directions: $  \{\text{Toward left}, \text{Toward right}\}  $. As illustrated in Fig.\ref{fig:invalid-gen}, the directed lighting configurations, (P: Right-side, D: Toward left) and (P: Left-side, D: Toward right), cannot be validly generalized to (P: Left-side, D: Toward left) or (P: Right-side, D: Toward right) setup, as this would leave the object in darkness. 
The above counterexample illustrates that \textbf{compositional generalization is not guaranteed for arbitrary factors and depends critically on how factors are defined.}

We acknowledge that defining factors is inherently challenging, as the physical world comprises a vast and highly entangled set of variables. Consequently, the goal is not to identify the single optimal factorization, but rather to select a set of proper factors that are both effective and practical. To this end, we propose the following several guiding principles for factor design:
\begin{itemize}
\item \textbf{Coverage:} The factor system should provide comprehensive coverage of the target space. Since our ultimate goal is to achieve strong generalization in real-world scenarios and our evaluation is conducted on a factor-based benchmark, any factor system that only partially represents the real world will lead to catastrophic failures in model performance. However, there exists a critical trade-off between coverage and granularity: overly fine-grained factors enhance representational fidelity at the cost of data efficiency, whereas overly coarse factors fail to span the full space, resulting in performance gaps.
\item \textbf{Experiment-driven:} The selection of factors is driven primarily by their empirical impact on model performance. In general, the most critical factors are those that have the strongest influence on overall success rate. For example, variations in object texture induce substantial shifts in the visual distribution and, in turn, significantly affect policy performance. In contrast, although changes in object color produce noticeable visual differences, they have only a limited effect on task success; we therefore exclude color from the object factor space.
The same empirical principle applies to compositional generalization. As shown in Fig.~\ref{fig:q1}, it is essential to verify that compositional generalization holds across the chosen factors before finalizing the representation. Our data collection strategy fundamentally relies on this property: should compositional generalization fail to hold, the iterative factor expansion would no longer ensure coverage of the full product space, making the entire approach ineffective.
\end{itemize}

\section{Experiments}
\label{sec:experiments}
We conducted extensive real-world experiments to validate the efficacy of \Name{}. Notably, we aim to address the following questions:
\begin{itemize}
    \item Does \Name{} suffice to achieve generalization over full product space? 
    \item How much does \Name{} accelerate the data flywheel compared to naive full-space iteration?
    \item How do we perform compositional generalization to improve data efficiency under a fixed budget?
    \item How does the growth of dimensionality affect the model performance?
\end{itemize}

\subsection{Experimental Setup}
\label{subsec:experimental setup}
\subsubsection{Model and Data.}
We select model with backbone inherits from \name{}~\cite{cheang2025gr3technicalreport}. Importantly, we keep the original backbone architecture intact, since all the performance gains stem from the proposed factor-wise iterative learning and compositional strategy. Furthermore, we train the VLA policy from scratch while freezing the backbone parameters during iterative learning, allowing us to isolate the effect of structured sampling.

We collect real-world manipulation demonstrations using ByteMini\cite{tian2025byte}. The collected trajectories involve grasping, placing tabletop objects, and hinge interactions such as opening and closing the door of the oven and microwave. Rather than passively aggregating demonstrations, data collection follows the process of \Name{}, where task instances are progressively expanded along selected factor dimensions across iterations. In addition, data collection is tightly coupled with the training iterations, where new compositions are selectively instantiated based on prior generalization performance. This strategy ensures that the demonstrations emphasize compositional coverage rather than exhaustive enumeration of the full compositional space.

\subsubsection{Factorize Demonstrations.}
\label{subsubsec:factorize data}
Based on the structured state representation of \Name{}, we factorize the experimental demonstrations into three dimensions: \textbf{Object ($\mathcal{O}$)-Action($\mathcal{A}$)-Environment($\mathcal{E}$)}. 
We acknowledge that the state representation adopted in \Name{} is neither unique nor necessarily optimal. Rather, it is designed to expose semantically controllable and physically meaningful variations in robotic manipulation tasks, building a reliable data flywheel and efficiently improving compositional generalization.

\paragraph{\textbf{\Name{}-Object}.}
(1)We consider three object-related factors: texture, geometry, and size. For Pick-and-Place tasks, objects are factorized along texture and geometry, while size is not explicitly considered due to constraints on gripper, as objects that are too large or too small cannot be reliably grasped. For Open-and-Close tasks, objects are factorized along texture (including transparency, specular reflection, and diffuse appearance) and size, whereas geometry is not explicitly factorized since hinge structures across appliances (e.g., microwave, oven, and drawer) are largely consistent. 
(2)We construct a $4 \times 4$ factor matrix for {Pick-and-Place} and a $4 \times 3$ factor matrix for {Open-and-Close} tasks. Alg.~\ref{alg:factor_object} is the applied to derive a minimal subset that allows generalization across the object space.

\paragraph{\textbf{\Name{}-Action}.}
(1)Action space is parameterized by planar position and rotation. Specifically, both skills fluctuate the planar position{$(x, y)$}, while the vertical dimension $z$ is not explicitly varied due to practical hardware constraints and stability considerations in real-world data collection. For rotation, we only vary the dimension of {yaw($y$)} while keeping other rotation axis fixed to maintain controlled and reproducible real-world execution.
(2)For Pick-and-Place, all three dimensions are retained, yielding a $4 \times 2 \times 3$ tensor. In contrast, for Open-and-Close, we consider a $4 \times 2$ action factor space. During exploration in $f(D_\mathcal{O})\mathcal{A}$, the ratio of object factor inherited from $f(D_\mathcal{O})$ is preserved.

\paragraph{\textbf{\Name{}-Environment}.}
(1)We introduce two factors in this space: Macro and Micro. For the Macro dimension, environments are factorized by shadow direction and color temperature, while light intensity is kept approximately constant to isolate other variations. At the Micro level, we randomly vary the surface material while keeping background clutter fixed. (2)The environment factor space forms a $3 \times 3$ matrix for both Pick-and-Place and Open-and-Close.

\subsubsection{Settings.}
During the real-world iterative process, we perform five tests for each subset in the evaluation set when assessing different skills. We measure an action is successful through applying skill-specific metrics, such as the grasp success rate for Pick-and-Place. Each test is assigned a score, and the final result is reported as the average score across the five runs. We evaluate the policy using region-based scoring. Only fully completed operations receive a score of 1, while partially completed or failed attempts receive a score of 0. This binary scoring scheme provides a rigorous and consistent metric for evaluating the compositional generalization of specified policy. Finally, the scores from all experimental settings are calculated to assess the performance of the current policy and guide the design of the next iteration if necessary. 
Notably, we consider below three groups to evaluate the improvement in data efficiency and the performance differences in compositional generalization introduced by \Name{}.

\begin{enumerate}
    \item 
    \textbf{\Name{}-Factors-Ratio group}: Based on Alg.~\ref{alg:factor_object}, constructs demonstrations by progressively increasing the coverage of task-relevant factors while controlling the ratio among factor spaces.
    \item
    \textbf{\Name{}-Factors-Mixture group}: Increases the overall number of demonstrations from a quasi-uniform distribution without explicitly accounting for the factor structure.
    \item
    \textbf{Gaussian group}: Samples demonstrations purely at random from a gaussian distribution without exploiting factor-aware data composition.
\end{enumerate}

\subsection{Main Results.}
\label{subsec:results}
%

\paragraph{Q1: Does \Name{} suffice to achieve generalization over full product space?}

Empirical results confirm that generalization achieved on the compact subset $f(D_\mathcal{O})\mathcal{A}$ transfers seamlessly to the entire product space $\mathcal{O}\mathcal{A}$. The resulting $D_\mathcal{OA}$ (obtained by Alg.~\ref{alg:factor_object}) satisfies
\begin{equation}
\mathrm{Orb}_{H_{f(D_\mathcal{O})\mathcal{A}}}\bigl(D_\mathcal{OA}\bigr) 
= f(D_\mathcal{O})\mathcal{A},
\end{equation}
which implies that the policy can generalize over $f(D_\mathcal{O})\mathcal{A}$. However, its generalization capability over the full $\mathcal{O}\mathcal{A}$ space remains unclear. Therefore, empirical evaluation is required to verify whether training on $D_\mathcal{OA}$ enables generalization to the complete $\mathcal{O}\mathcal{A}$ space.

\begin{figure}
    \centering
    \begin{subfigure}[t]{0.42\linewidth}
        \centering
        \includegraphics[width=\linewidth]{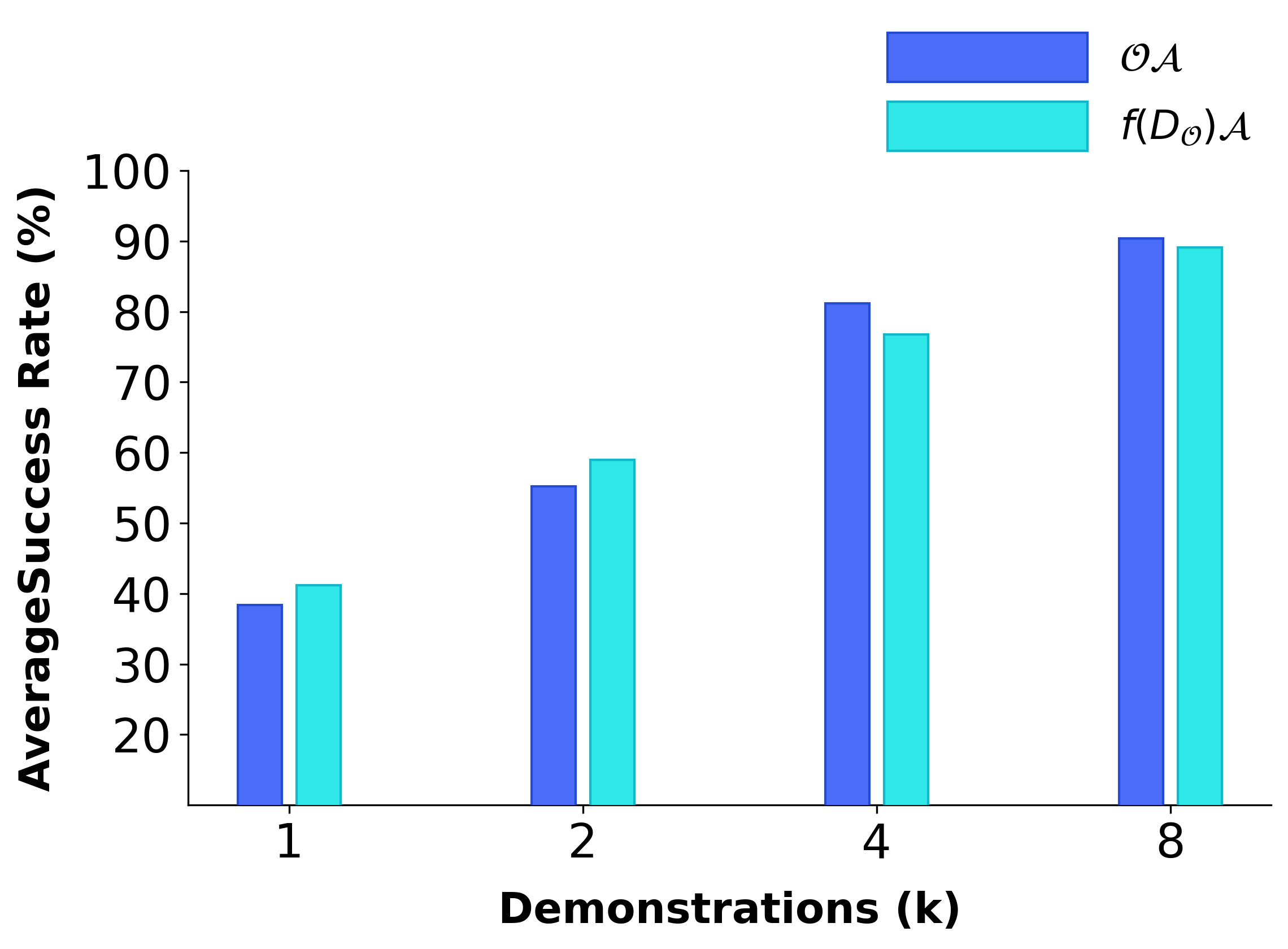}
    \end{subfigure}
    \hfill
    \begin{subfigure}[t]{0.57\linewidth}
        \centering
        \includegraphics[width=\linewidth]{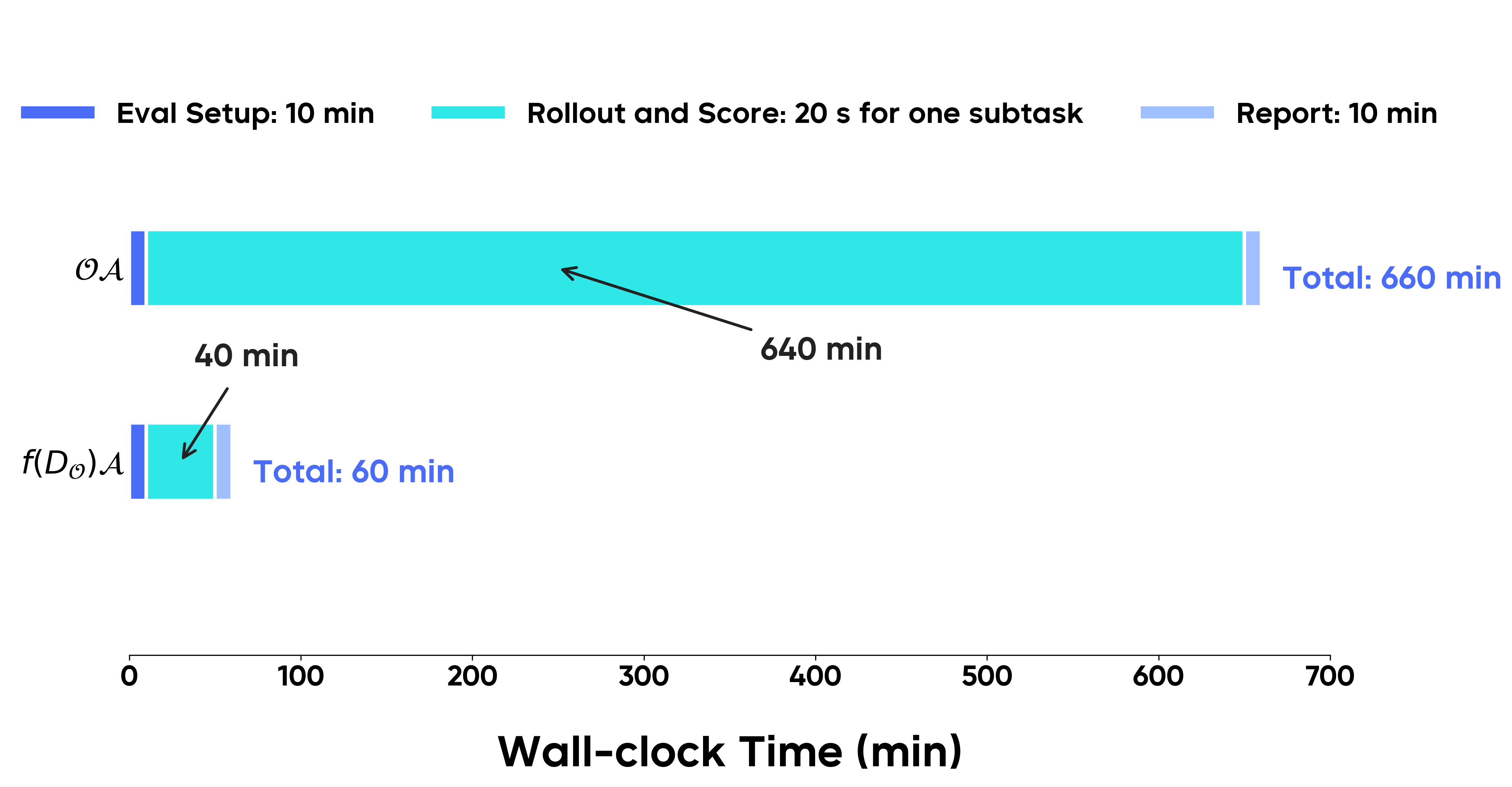}
    \end{subfigure}
    \caption{
        \textbf{Generalization evaluation on $f(D_\mathcal{O})\mathcal{A}$ and $\mathcal{O}\mathcal{A}$}.
        (left) The overall success rates evaluated on $f(D_\mathcal{O})\mathcal{A}$ and on the complete $\mathcal{O}\mathcal{A}$ space exhibit only a slight difference. (right) Time needed for one iteration of the data flywheel between entire $\mathcal{OA}$ and reduced $f(D_{\mathcal{O}})\mathcal{A}$.
    }
    \label{fig:o-oa}
\end{figure}

As shown in Fig.~\ref{fig:o-oa} (left), as the data volume increases, the evaluation gap between 
$f(D_\mathcal{O})\mathcal{A}$ and the complete $\mathcal{O}\mathcal{A}$ gradually diminishes. 
This empirical trend suggests that:
\begin{equation}
\mathrm{Orb}_{H_{\mathcal{O}\mathcal{A}}}\bigl(D_\mathcal{OA}\bigr)
= \mathcal{O}\mathcal{A}.
\end{equation}

In other words, once a subset achieves generalization within the restricted space 
$f(D_\mathcal{O})\mathcal{A}$, it tends to extend to the full $\mathcal{O}\mathcal{A}$ space. This finding further indicates that once the model achieves robust generalization within the compact subset, there is a strong likelihood that this generalization extends to the entire factor space.

\paragraph{Q2: How much does \Name{} accelerate the data flywheel compared to naive full-space iteration?}

We compare the time cost of a single iteration of the data flywheel under the full product space $  \mathcal{OA}  $ and the reduced space $  f(D_{\mathcal{O}})\mathcal{A}  $ in the right panel of Fig.~\ref{fig:o-oa}.
For the Pick-and-Place task in the full $  \mathcal{OA}  $ space, one complete iteration requires evaluating all possible factor combinations, resulting in $  4 \times 2 \times 3 \times 4 \times 4 = 384  $ possibilities and $384 \times 5=1920$ evaluations.
In the reduced space $  f(D_{\mathcal{O}}) \mathcal{A}  $, the object factor set is much smaller ($  |f(D_{\mathcal{O}})| = 7  $, as shown in iteration 4 of Fig.~\ref{fig:iteration}), reducing the number of possibilities to $  4 \times 2 \times 3 \times 7 = 168  $. While this already represents a meaningful reduction, the cost remains substantial.
To further lower the rollout budget without sacrificing coverage, we adopt a random sampling strategy guided by the factor ratios in $  D_{\mathcal{O}}  $. Specifically, for each evaluation, we sample the object factor composition according to the factor ratio within $  f(D_{\mathcal{O}})  $. As a result, the rollout count drops dramatically to just $  4 \times 2 \times 3  \times 5 = 120  $, $16\times$ less than original number. \textbf{This reduction dramatically accelerates the data flywheel by 16×}, transforming what was once prohibitively expensive into a highly efficient and scalable iteration cycle.

\begin{figure}[!t]
    \centering
    \includegraphics[width=\linewidth]{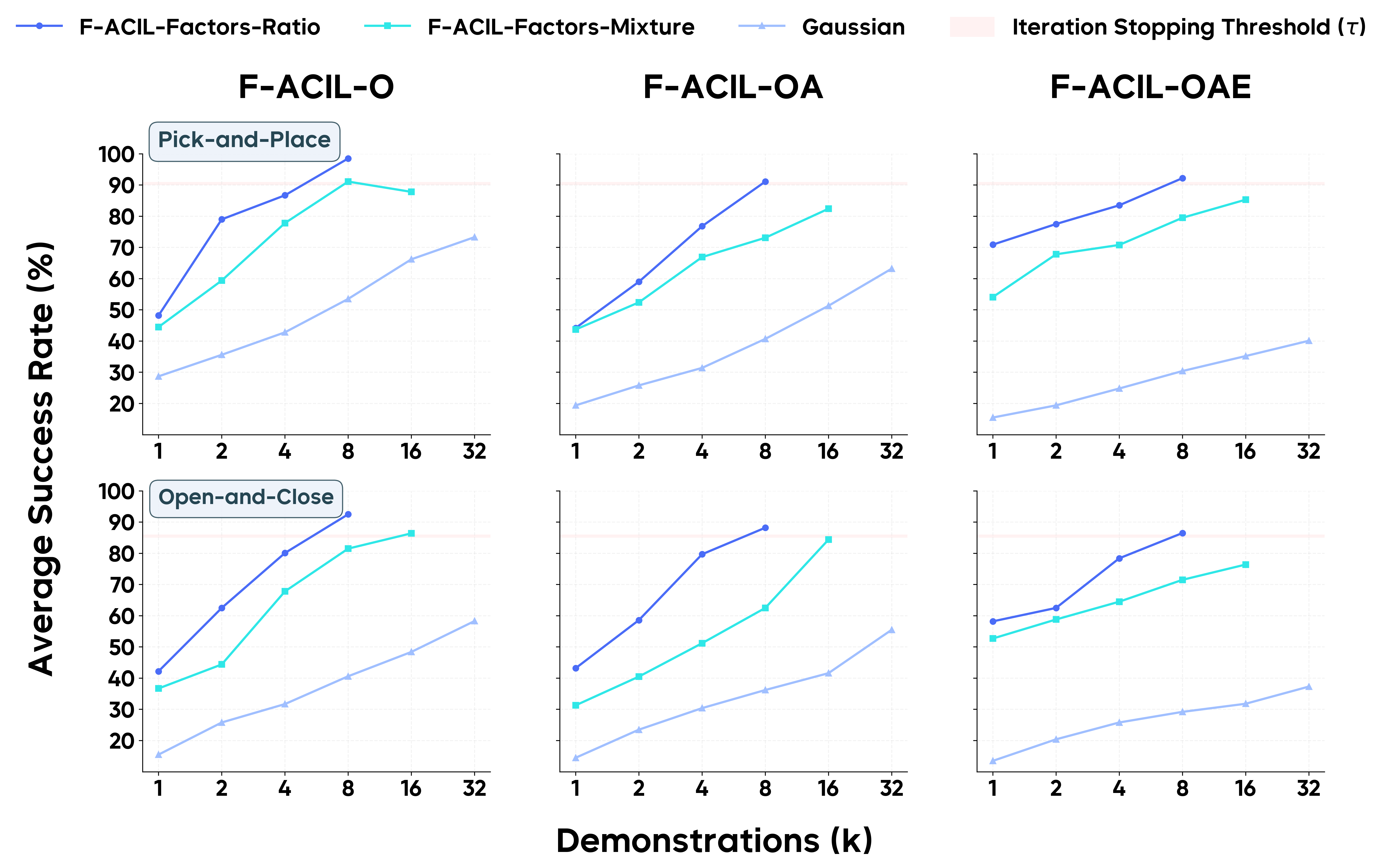}
    \caption{
        \textbf{Comparison of data strategies across factor spaces.}
        Each curve corresponds to a different data collection strategy used. \Name{}-O, -OA and -OAE represent the iterative subset search under different factor spaces.
    }
    \label{fig:comparative_analysis}
\end{figure}

\paragraph{Q3: How do we perform compositional generalization to improve data efficiency under a fixed budget?}




We study the generalization of \Name{} in three factorized spaces: object, action, and environment.
As illustrated in Fig.~\ref{fig:comparative_analysis}, there is a consistent trend that the proposed \Name{}-Factors-Ratio strategy achieves comparable performance with substantially fewer demonstrations across both skill groups (Pick-and-Place and Open-and-Close).

In Pick-and-Place experiments within the object factor space, \Name{}-Factors-Ratio achieves around 80–90\% success rate with fewer than 4k trajectories for both skills. In contrast, \Name{}-Factors-Mixture typically requires approximately 8k demonstrations to reach a comparable performance level, corresponding to an approximate 2–3× reduction in data requirement.
This advantage persists as the factor dimensionality increases. In the object–action setting of Open-and-Close, \Name{}-Factors-Ratio exceeds 80\% success with 4k demonstrations and further improves to 85\% with 8k, whereas \Name{}-Factors-Mixture requires roughly 16k demonstrations to approach a similar threshold, indicating a 3–4× gain in data efficiency.
In the most challenging object–action–environment factor space, both skills trained with \Name{}-Factors-Ratio achieve approximately 85–95\% success using 6–8k demonstrations. Notably, factor-aware strategies already provide substantial gains in the low-data regime (e.g., 1k demonstrations of \Name{}-Factor-Ratio and \Name{}-Factor-Mixture), indicating that environmental variations can be transferred more effectively once object and action factors have been factorized and generalized through iterative learning.
Compared to these structured factor-wise strategies, the Gaussian baseline demonstrates the worst generalization, with performance improving slowly only as much more demonstrations are introduced. In practice, it often requires over 32k demonstrations to reach performance levels that \Name{}-Factors-Ratio achieves with 2-4k samples, corresponding to at least 5–10× increments in data efficiency. 
Furthermore, the comparison reveals that strategy which overlook rational factor design and appropriate data ratios tend to require dramatically larger datasets before the expected generalization could emerge.

Despite these trends, the two skill groups also exhibit distinct behaviors. 
For Pick-and-Place, performance improves more smoothly as the amount of demonstrations increases, and the gap between the \Name{}-Factors-Ratio and \Name{}-Factors-Mixture remains relatively consistent across different factor spaces. This trend suggests that compositional coverage mainly facilitates generalization over object-centric variations of the simple skill.
However, Open-and-Close experiment exhibits substantially larger performance gaps in the object–action factor space. Typically, the failure rate of \Name{}-Factor-Mixture remains high when the number of demonstrations is below 8k, whereas the \Name{}-Factor-Ratio achieves better performance. This suggests that skills involving articulated interactions are more sensitive to variations along the action-related components, making them more vulnerable to distribution shifts between training and evaluation. Consequently, careful distribution design with appropriate factor-wise ratio adjustment is critical for improving generalization.

\begin{table}[t]
\centering
\caption{Changes in success rate with changes in dimensions}
\label{tab:suc_rate}
\begin{tabular}{cc 
  >{\columncolor{SkyBlue!50}}c 
  >{\columncolor{SkyBlue!30}}c 
  >{\columncolor{SkyBlue!10}}c 
  >{\columncolor{SkyBlue!5}}c 
  >{\columncolor{SkyBlue!0}}c}
\toprule
Task & Benchmark & Gaussian-32k & Gaussian-16k & Gaussian-8k & Gaussian-4k & Gaussian-2k \\
\midrule
\multirow{3}{*}{Pick-and-Place}
& \Name{}-O & \textbf{73.3} & 69.2 & 63.3 & 51.4 & 39.8 \\
& \Name{}-OA & \textbf{69.2} & 54.3 & 41.7 & 35.4 & 22.8 \\
& \Name{}-OAE & \textbf{40.1} & 35.2 & 30.4 & 24.8 & 19.4 \\
\midrule
\multirow{3}{*}{Open-and-Close}
& \Name{}-O & \textbf{58.3} & 48.4 & 40.6 & 31.7 & 25.8 \\
& \Name{}-OA & \textbf{55.5} & 41.6 & 36.2 & 30.4 & 23.5 \\
& \Name{}-OAE & \textbf{37.3} & 31.8 & 29.2 & 25.8 & 20.4 \\
\bottomrule
\end{tabular}
\end{table}

\paragraph{Q4: How does the growth of dimensionality affect the model performance?}

A common belief in the robotic learning community is that simply increasing dataset size will reliably lead to higher performance, as predicted by scaling laws. This intuition is not entirely unfounded: empirical evidence consistently shows exponential improvements in model performance with scale. The results in Tab.~\ref{tab:suc_rate} provide a concrete proof of this pattern: performance improves monotonically with dataset size.
However, the slope (i.e., the scaling exponent $\alpha$ in the scaling relation $L \approx N^{-\alpha}$ ) of these power laws can vary dramatically depending on the dimensionality of the data manifold or task space. Blindly scaling up dataset volume without accounting for dimensionality often results in substantially diminished returns — a manifestation of the curse of dimensionality in high-dimensional regimes.

\begin{figure}
    \centering
    \includegraphics[width=\linewidth]{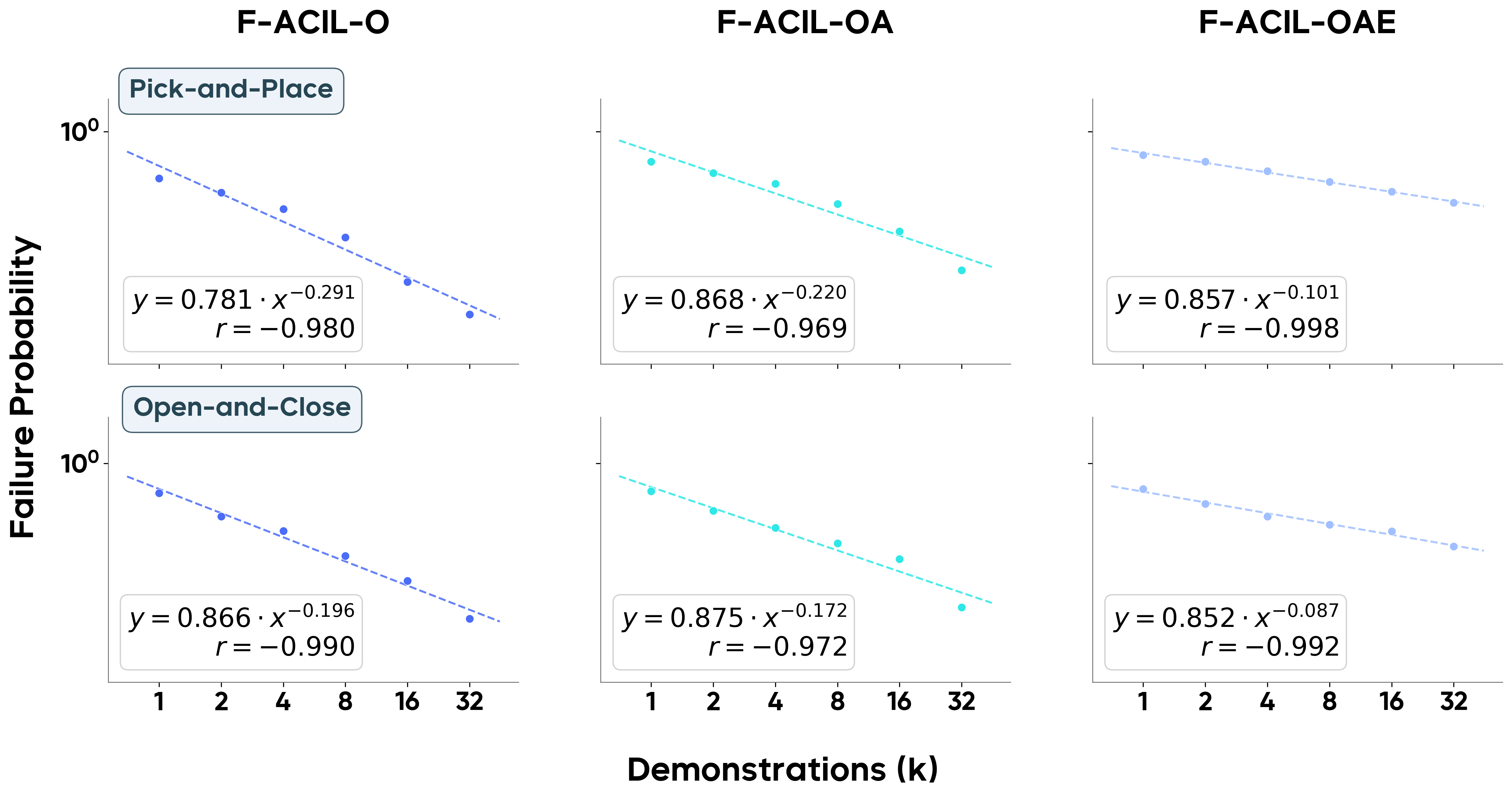}

    \caption{
        \textbf{Scaling Pattern w/o \Name{} under Different Benchmark}
        {
         We compare the power-law fits under different benchmark. Each benchmark reflects the model's generalization ability over the corresponding factor space: $\mathcal{O}$ (object), $\mathcal{OA}$ (object–action), and $\mathcal{OAE}$ (full space).
        }
    }
    \label{fig:pure_random_pattern}
\end{figure}

As shown in Fig.~\ref{fig:pure_random_pattern}: 
\begin{itemize}
    \item In low-intrinsic-dimensional regimes, scaling exponents tend to be smaller, yielding faster and more efficient performance gains from additional data. For the Pick-and-Place task, the fitted scaling exponent reaches to -0.291 when evaluated on the object-only benchmark $\mathcal{O}$. Similarly, for the Open-and-Close task, the exponent under the $\mathcal{O}$ benchmark is -0.196.
    \item In higher-dimensional regimes, the scaling exponent increases, acquiring more data exponentially to achieve comparable improvements. For Pick-and-Place, the exponent progressively increase from -0.291 (on $\mathcal{O}$) to -0.220 (on $\mathcal{OA}$) and further to -0.101 (on the full $\mathcal{OAE}$ space). A qualitatively similar trend holds for Open-and-Close, where the exponent rises to -0.172 under $\mathcal{OA}$ and -0.087 under $\mathcal{OAE}$.
\end{itemize}

\textbf{While scaling laws provide a principled foundation expecting gains from larger datasets, their practical efficacy highly depend on the underlying dimensionality of the data.} Effective strategies should therefore incorporate dimensionality-aware design rather than relying solely on data volume expansion.



\section{Limitations \& Conclusions}
\label{sec:conclusions}
\paragraph{Limitations and future work.} 
The variety of factor space and depth of factor dimensions can be extended further. In this work, we are unable to cover some promising factors space like \textbf{language} and \textbf{embodiment} due to limited time and resources. For factor dimensions such as object geometry, they can be further decomposed to align with the state space of the robot policy. In addition, while we assume factors are independent to each other in our formulation, it is recognized that factors like objects, actions, and environments could be subtly interleaved, offering opportunities for more sophisticated factorization.

An efficient way to explore more factor spaces and potential compositions is through robot simulation~\cite{mittal2025isaaclab,Xiang_2020_SAPIEN}. Comparing to real-world experiments mentioned previously, simulation experiment allows quantitative factor configuration and more diverse setups. For example, instead of small, medium, and large, the size of objects can be categorized by the volumes of the bounding box in the simulation. Besides, by combining data generation methods and automatic evaluation, simulation can largely accelerate data flywheel.

Looking ahead, the effective application of \Name{} in large-scale, real-world settings requires further investigation. Our current formulation assumes structured factor spaces where object, action, and environment can be reasonably enumerated and discretized. However, real-world robotic systems often involve additional complexities, including long-horizon task composition, dynamic environments, and evolving object distributions. Addressing these challenges may require extensions such as hierarchical factorization, adaptive refinement, and real-time discovery, which could broaden the applicability of \Name{} beyond experimental scenarios.

\paragraph{Conclusions.} In this work, we present \Name{} to improve data efficiency in robot policy learning with factorized state representations and sequential factor-wise exploration. By decomposing the high-dimensional state space into object, action, and environment factor spaces, \Name{} enables more systematic data collection. 
To further reduce data requirements while maintaining coverage of the full factor space, \Name{} leverages compositional generalization across factors, allowing models trained on sparse factor combinations to generalize broadly. 
Compared to Gaussian baselines, \Name{} achieves more than 45\% performance gain with 5–10$\times$ fewer demonstrations. 
Overall, our results highlight the importance of structured factorization and composition for building more efficient and generalizable data flywheel for robotic learning.

\clearpage
\section{Acknowledgements}
\label{sec:contributions}

We appreciate Chilam Cheang and Wanli Peng for their contributions during the early stages of the project, particularly in brainstorming and material preparation.

We thank Pengfei Zhang, Yongpeng Yang, Xin Zhao, Mingyang Wang, Yang Zhao, Shuzhai Guo, Juyi Bai, and Tianxiang Gong, Ziye Liu, and Xiaodong Lan for their efforts in data collection.

We thank Jingchao Qiao, Liqun Huang, Zhongren Cui, Jiawen Tian, Zhigang Han, Niu Hao, Liwei Zheng, and Zeyu Ren for their consistent maintenance of the hardware system.

We thank Xiao Ma, Zhengbang Zhu, Zetian Li, Jinming Guo, Yage Zhu, Yanhui Wang, Peirong Zhang, Chengyuan Luo, Degong Yang, Yang Liu, and Wenquan Dong for their assistance of the project.

\clearpage

\bibliographystyle{plainnat}
\bibliography{main}

\begin{thebibliography}{47}
\providecommand{\natexlab}[1]{#1}
\providecommand{\url}[1]{\texttt{#1}}
\expandafter\ifx\csname urlstyle\endcsname\relax
  \providecommand{\doi}[1]{doi: #1}\else
  \providecommand{\doi}{doi: \begingroup \urlstyle{rm}\Url}\fi

\bibitem[Brohan et~al.(2023)Brohan, Brown, Carbajal, Chebotar, Chen, Choromanski, Ding, Driess, Dubey, Finn, et~al.]{brohan2023rt}
Anthony Brohan, Noah Brown, Justice Carbajal, Yevgen Chebotar, Xi~Chen, Krzysztof Choromanski, Tianli Ding, Danny Driess, Avinava Dubey, Chelsea Finn, et~al.
\newblock Rt-2: Vision-language-action models transfer web knowledge to robotic control.
\newblock \emph{arXiv preprint arXiv:2307.15818}, 2023.

\bibitem[Cheang et~al.(2025)Cheang, Chen, Cui, Hu, Huang, Kong, Li, Li, Liu, Ma, et~al.]{cheang2025gr3technicalreport}
Chilam Cheang, Sijin Chen, Zhongren Cui, Yingdong Hu, Liqun Huang, Tao Kong, Hang Li, Yifeng Li, Yuxiao Liu, Xiao Ma, et~al.
\newblock Gr-3 technical report.
\newblock \emph{arXiv preprint at arXiv:2507.15493}, 2025.

\bibitem[Chen et~al.(2023)Chen, Djolonga, Padlewski, Mustafa, Changpinyo, Wu, Ruiz, Goodman, Wang, Tay, et~al.]{chen2023pali}
Xi~Chen, Josip Djolonga, Piotr Padlewski, Basil Mustafa, Soravit Changpinyo, Jialin Wu, Carlos~Riquelme Ruiz, Sebastian Goodman, Xiao Wang, Yi~Tay, et~al.
\newblock Pali-x: On scaling up a multilingual vision and language model.
\newblock \emph{arXiv preprint arXiv:2305.18565}, 2023.

\bibitem[Chi et~al.(2024)Chi, Xu, Feng, Cousineau, Du, Burchfiel, Tedrake, and Song]{chi2024diffusionpolicy}
Cheng Chi, Zhenjia Xu, Siyuan Feng, Eric Cousineau, Yilun Du, Benjamin Burchfiel, Russ Tedrake, and Shuran Song.
\newblock Diffusion policy: Visuomotor policy learning via action diffusion.
\newblock \emph{The International Journal of Robotics Research}, 2024.

\bibitem[Driess et~al.(2023)Driess, Huang, Li, Tedrake, and Toussaint]{driess2023Learning}
Danny Driess, Zhiao Huang, Yunzhu Li, Russ Tedrake, and Marc Toussaint.
\newblock Learning multi-object dynamics with compositional neural radiance fields.
\newblock In \emph{Proceedings of The 6th Conference on Robot Learning}, pages 1755--1768, 2023.

\bibitem[Ebert et~al.(2021)Ebert, Yang, Schmeckpeper, Bucher, Georgakis, Daniilidis, Finn, and Levine]{ebert2021bridge}
Frederik Ebert, Yanlai Yang, Karl Schmeckpeper, Bernadette Bucher, Georgios Georgakis, Kostas Daniilidis, Chelsea Finn, and Sergey Levine.
\newblock Bridge data: Boosting generalization of robotic skills with cross-domain datasets.
\newblock \emph{arXiv preprint arXiv:2109.13396}, 2021.

\bibitem[Fang et~al.(2025)Fang, Zhang, Tong, and Feng]{fang2025intentionexecutionprobinggeneralization}
Irving Fang, Juexiao Zhang, Shengbang Tong, and Chen Feng.
\newblock From intention to execution: Probing the generalization boundaries of vision-language-action models.
\newblock \emph{arXiv preprint arXiv:2506.09930}, 2025.

\bibitem[Fei et~al.(2025)Fei, Wang, Shi, Dai, Cai, Qian, Ji, He, Zhang, Fei, et~al.]{fei2025liberoplus}
Senyu Fei, Siyin Wang, Junhao Shi, Zihao Dai, Jikun Cai, Pengfang Qian, Li~Ji, Xinzhe He, Shiduo Zhang, Zhaoye Fei, et~al.
\newblock Libero-plus: In-depth robustness analysis of vision-language-action models.
\newblock \emph{arXiv preprint arXiv:2510.13626}, 2025.

\bibitem[Gao et~al.(2024)Gao, Xie, Xiao, Finn, and Sadigh]{gao2024efficient}
Jensen Gao, Annie Xie, Ted Xiao, Chelsea Finn, and Dorsa Sadigh.
\newblock Efficient data collection for robotic manipulation via compositional generalization.
\newblock \emph{arXiv preprint at arXiv:2403.05110}, 2024.

\bibitem[Gao et~al.(2026)Gao, Belkhale, Dasari, Balakrishna, Shah, and Sadigh]{gao2026taxonomy}
Jensen Gao, Suneel Belkhale, Sudeep Dasari, Ashwin Balakrishna, Dhruv Shah, and Dorsa Sadigh.
\newblock A taxonomy for evaluating generalist robot manipulation policies.
\newblock \emph{arXiv preprint at arXiv:2503.01238v3}, 2026.

\bibitem[Guruprasad et~al.(2024)Guruprasad, Sikka, Song, Wang, and Liang]{guruprasad2024benchmarkingvisionlanguage}
Pranav Guruprasad, Harshvardhan Sikka, Jaewoo Song, Yangyue Wang, and Paul~Pu Liang.
\newblock Benchmarking vision, language, \& action models on robotic learning tasks.
\newblock \emph{arXiv preprint arXiv:2411.05821}, 2024.

\bibitem[Hansen and Wang(2021)]{hansen2021generalizationrlsoft}
Nicklas Hansen and Xiaolong Wang.
\newblock Generalization in reinforcement learning by soft data augmentation.
\newblock \emph{arXiv preprint at arXiv:2011.13389}, 2021.

\bibitem[Hu et~al.(2025)Hu, Lin, Sheng, Wen, You, and Gao]{hu2025datascalinglaws}
Yingdong Hu, Fanqi Lin, Pingyue Sheng, Chuan Wen, Jiacheng You, and Yang Gao.
\newblock Data scaling laws in imitation learning for robotic manipulation.
\newblock \emph{arXiv preprint arXiv:2410.18647}, 2025.

\bibitem[Huang et~al.(2025)Huang, Liao, Feng, Jiang, Liu, Li, Yao, and Ren]{huang2025adc}
Siyuan Huang, Yue Liao, Siyuan Feng, Shu Jiang, Si~Liu, Hongsheng Li, Maoqing Yao, and Guanghui Ren.
\newblock Adversarial data collection: Human-collaborative perturbations for efficient and robust robotic imitation learning.
\newblock \emph{arXiv preprint at arXiv:2503.11646}, 2025.

\bibitem[Ito et~al.(2022)Ito, Klinger, Schultz, Murray, Cole, and Rigotti]{ito2022compositionalgeneralizationabstractrepresentations}
Takuya Ito, Tim Klinger, Douglas~H. Schultz, John~D. Murray, Michael~W. Cole, and Mattia Rigotti.
\newblock Compositional generalization through abstract representations in human and artificial neural networks.
\newblock \emph{arXiv preprint arXiv:2209.07431}, 2022.

\bibitem[James et~al.(2020)James, Ma, Arrojo, and Davison]{james2020rlbench}
Stephen James, Zicong Ma, David~Rovick Arrojo, and Andrew~J Davison.
\newblock Rlbench: The robot learning benchmark \& learning environment.
\newblock \emph{IEEE Robotics and Automation Letters}, 5\penalty0 (2):\penalty0 3019--3026, 2020.

\bibitem[Kang et~al.(2023)Kang, Ma, Du, Pang, and Yan]{kang2023efficient}
Bingyi Kang, Xiao Ma, Chao Du, Tianyu Pang, and Shuicheng Yan.
\newblock Efficient diffusion policies for offline reinforcement learning.
\newblock \emph{Advances in Neural Information Processing Systems}, 36:\penalty0 67195--67212, 2023.

\bibitem[Li et~al.(2024)Li, Jing, Li, Zhai, Wu, and Jia]{li2024context}
Chuanhao Li, Chenchen Jing, Zhen Li, Mingliang Zhai, Yuwei Wu, and Yunde Jia.
\newblock In-context compositional generalization for large vision-language models.
\newblock In \emph{Proceedings of the 2024 Conference on Empirical Methods in Natural Language Processing}, pages 17954--17966, 2024.

\bibitem[Li et~al.(2025{\natexlab{a}})Li, Song, Zhao, Wang, Ding, Wang, Zeng, and Li]{li2025spatialforce}
Fuhao Li, Wenxuan Song, Han Zhao, Jingbo Wang, Pengxiang Ding, Donglin Wang, Long Zeng, and Haoang Li.
\newblock Spatial forcing: Implicit spatial representation alignment for vision-language-action model.
\newblock \emph{arXiv preprint at arXiv:2510.12276}, 2025{\natexlab{a}}.

\bibitem[Li et~al.(2025{\natexlab{b}})Li, Zhang, Rao, and Cheng]{li2025unveil}
Tianle Li, Jihai Zhang, Yongming Rao, and Yu~Cheng.
\newblock Unveiling the compositional ability gap in vision-language reasoning model.
\newblock \emph{arXiv preprint at arXiv:2505.19406}, 2025{\natexlab{b}}.

\bibitem[Li et~al.(2023)Li, Liu, Zhang, Yu, Xu, Wu, Cheang, Jing, Zhang, Liu, et~al.]{li2023vision}
Xinghang Li, Minghuan Liu, Hanbo Zhang, Cunjun Yu, Jie Xu, Hongtao Wu, Chilam Cheang, Ya~Jing, Weinan Zhang, Huaping Liu, et~al.
\newblock Vision-language foundation models as effective robot imitators.
\newblock \emph{arXiv preprint arXiv:2311.01378}, 2023.

\bibitem[Li(2025)]{li2025theoreticalanalysiscompositionalgeneralization}
Yuanpeng Li.
\newblock A theoretical analysis of compositional generalization in neural networks: A necessary and sufficient condition.
\newblock \emph{arXiv preprint arXiv:2505.02627}, 2025.

\bibitem[Liu et~al.(2023)Liu, Zhu, Gao, Feng, Liu, Zhu, and Stone]{liu2023liberobenchmarkingknowledgetransfer}
Bo~Liu, Yifeng Zhu, Chongkai Gao, Yihao Feng, Qiang Liu, Yuke Zhu, and Peter Stone.
\newblock Libero: Benchmarking knowledge transfer for lifelong robot learning.
\newblock \emph{arXiv preprint at arXiv:2306.03310}, 2023.

\bibitem[Lu et~al.(2025)Lu, Guo, Zhang, Zhou, Jiang, Gao, Tang, and Wang]{lu2025vlarl}
Guanxing Lu, Wenkai Guo, Chubin Zhang, Yuheng Zhou, Haonan Jiang, Zifeng Gao, Yansong Tang, and Ziwei Wang.
\newblock Vla-rl: Towards masterful and general robotic manipulation with scalable reinforcement learning.
\newblock \emph{arXiv preprint at arXiv:2505.18719}, 2025.

\bibitem[Lécuyer et~al.(2022)Lécuyer, Jachiet, Magnien, and Tabourier]{lecuyer2022tailoredvertexorderingfaster}
Fabrice Lécuyer, Louis Jachiet, Clémence Magnien, and Lionel Tabourier.
\newblock Tailored vertex ordering for faster triangle listing in large graphs.
\newblock \emph{arXiv preprint arXiv:2203.04774}, 2022.

\bibitem[Mendez et~al.(2022)Mendez, Hussing, Gummadi, and Eaton]{mendez2022composuite}
Jorge~A. Mendez, Marcel Hussing, Meghna Gummadi, and Eric Eaton.
\newblock Composuite: A compositional reinforcement learning benchmark.
\newblock \emph{arXiv preprint at arXiv:2207.04136}, 2022.

\bibitem[Mittal et~al.(2025)Mittal, Roth, Tigue, Richard, Zhang, Du, Serrano-Muñoz, Yao, Zurbrügg, Rudin, Wawrzyniak, Rakhsha, Denzler, Heiden, Borovicka, Ahmed, Akinola, Anwar, et~al.]{mittal2025isaaclab}
Mayank Mittal, Pascal Roth, James Tigue, Antoine Richard, Octi Zhang, Peter Du, Antonio Serrano-Muñoz, Xinjie Yao, René Zurbrügg, Nikita Rudin, Lukasz Wawrzyniak, Milad Rakhsha, Alain Denzler, Eric Heiden, Ales Borovicka, Ossama Ahmed, Iretiayo Akinola, Abrar Anwar, et~al.
\newblock Isaac lab: A gpu-accelerated simulation framework for multi-modal robot learning.
\newblock \emph{arXiv preprint arXiv:2511.04831}, 2025.
\newblock URL \url{https://arxiv.org/abs/2511.04831}.

\bibitem[Nachum et~al.(2018)Nachum, Gu, Lee, and Levine]{nachum2018data}
Ofir Nachum, Shixiang Gu, Honglak Lee, and Sergey Levine.
\newblock Data-efficient hierarchical reinforcement learning.
\newblock \emph{arXiv preprint at arXiv:1805.08296}, 2018.

\bibitem[Qu et~al.(2025)Qu, Song, Chen, Yao, Ye, Ding, Wang, Gu, Zhao, Wang, et~al.]{qu2025spatialvla}
Delin Qu, Haoming Song, Qizhi Chen, Yuanqi Yao, Xinyi Ye, Yan Ding, Zhigang Wang, JiaYuan Gu, Bin Zhao, Dong Wang, et~al.
\newblock Spatialvla: Exploring spatial representations for visual-language-action model.
\newblock \emph{arXiv preprint arXiv:2501.15830}, 2025.

\bibitem[Ren et~al.(2024)Ren, Cong, Chen, and Long]{ren2024learning}
Yu~Ren, Yang Cong, Ronghan Chen, and Jiahao Long.
\newblock Learning generalizable 3d manipulation with 10 demonstrations.
\newblock \emph{arXiv preprint arXiv:2411.10203}, 2024.

\bibitem[Sedlacek et~al.(2025)Sedlacek, Yefanov, Ponimatkin, Bardhan, Pilc, Fourmy, Kazakos, Snoek, Sivic, and Petrik]{sedlacek2025realmrealtosimvalidatedbenchmark}
Martin Sedlacek, Pavlo Yefanov, Georgy Ponimatkin, Jai Bardhan, Simon Pilc, Mederic Fourmy, Evangelos Kazakos, Cees G.~M. Snoek, Josef Sivic, and Vladimir Petrik.
\newblock Realm: A real-to-sim validated benchmark for generalization in robotic manipulation.
\newblock \emph{arXiv preprint arXiv:2512.19562}, 2025.

\bibitem[Tian et~al.(2025)Tian, Huang, Cui, Qiao, Xu, Ma, and Ren]{tian2025byte}
Jiawen Tian, Liqun Huang, Zhongren Cui, Jingchao Qiao, Jiafeng Xu, Xiao Ma, and Zeyu Ren.
\newblock Bytewrist: A parallel robotic wrist enabling flexible and anthropomorphic motion for confined spaces.
\newblock \emph{arXiv preprint at arXiv:2509.18084}, 2025.

\bibitem[Vijayaraghavan et~al.(2024)Vijayaraghavan, Queisser, Flores, and Tani]{vijayaraghavan2024development}
Prasanna Vijayaraghavan, Jeffrey~Frederic Queisser, Sergio~Verduzco Flores, and Jun Tani.
\newblock Development of compositionality and generalization through interactive learning of language and action of robots.
\newblock \emph{arXiv preprint at arXiv:2403.19995}, 2024.

\bibitem[Wang et~al.(2024)Wang, Zhao, Du, Adelson, and Tedrake]{wang2024poco}
Lirui Wang, Jialiang Zhao, Yilun Du, Edward~H. Adelson, and Russ Tedrake.
\newblock Poco: Policy composition from and for heterogeneous robot learning.
\newblock \emph{arXiv preprint at arXiv:2402.02511}, 2024.

\bibitem[Wang et~al.(2023)Wang, Mao, Hsu, Zhao, Wu, and Gao]{wang2023program}
Renhao Wang, Jiayuan Mao, Joy Hsu, Hang Zhao, Jiajun Wu, and Yang Gao.
\newblock Programmatically grounded, compositionally generalizable robotic manipulation.
\newblock \emph{arXiv preprint at arXiv:2304.13826}, 2023.

\bibitem[Wang et~al.(2025)Wang, You, Hu, Li, and Gao]{wang2025skil}
Shengjie Wang, Jiacheng You, Yihang Hu, Jiongye Li, and Yang Gao.
\newblock Skil: Semantic keypoint imitation learning for generalizable data-efficient manipulation.
\newblock \emph{arXiv preprint arXiv:2501.14400}, 2025.

\bibitem[Wen et~al.(2024)Wen, Zhu, Li, Zhu, Wu, Xu, Liu, Cheng, Shen, Peng, et~al.]{wen2024tinyvla}
Junjie Wen, Yichen Zhu, Jinming Li, Minjie Zhu, Kun Wu, Zhiyuan Xu, Ning Liu, Ran Cheng, Chaomin Shen, Yaxin Peng, et~al.
\newblock Tinyvla: Towards fast, data-efficient vision-language-action models for robotic manipulation.
\newblock \emph{arXiv preprint arXiv:2409.12514}, 2024.

\bibitem[Wu et~al.(2022)Wu, Xie, Lian, Wang, Guo, Chen, Schaal, and Tomizuka]{wu2022zeroshot}
Zheng Wu, Yichen Xie, Wenzhao Lian, Changhao Wang, Yanjiang Guo, Jianyu Chen, Stefan Schaal, and Masayoshi Tomizuka.
\newblock Zero-shot policy transfer with disentangled task representation of meta-reinforcement learning.
\newblock \emph{arXiv preprint at arXiv:2210.00350}, 2022.

\bibitem[Xiang et~al.(2020)Xiang, Qin, Mo, Xia, Zhu, Liu, Liu, Jiang, Yuan, Wang, Yi, Chang, Guibas, and Su]{Xiang_2020_SAPIEN}
Fanbo Xiang, Yuzhe Qin, Kaichun Mo, Yikuan Xia, Hao Zhu, Fangchen Liu, Minghua Liu, Hanxiao Jiang, Yifu Yuan, He~Wang, Li~Yi, Angel~X. Chang, Leonidas~J. Guibas, and Hao Su.
\newblock {SAPIEN}: A simulated part-based interactive environment.
\newblock In \emph{The IEEE Conference on Computer Vision and Pattern Recognition (CVPR)}, June 2020.

\bibitem[Xu et~al.(2025)Xu, Mao, Du, Lozáno-Pérez, Kaelbling, and Hsu]{xu2025setup}
Yiqing Xu, Jiayuan Mao, Yilun Du, Tomas Lozáno-Pérez, Leslie~Pack Kaelbling, and David Hsu.
\newblock "set it up!": Functional object arrangement with compositional generative models.
\newblock \emph{arXiv preprint arXiv:2405.11928}, 2025.

\bibitem[Ye et~al.(2025)Ye, Zhang, Wang, Wang, Zhang, and Zhu]{ye2025vlar1}
Angen Ye, Zeyu Zhang, Boyuan Wang, Xiaofeng Wang, Dapeng Zhang, and Zheng Zhu.
\newblock Vla-r1: Enhancing reasoning in vision-language-action models.
\newblock \emph{arXiv preprint at arXiv:2510.01623}, 2025.

\bibitem[Zhang et~al.(2025{\natexlab{a}})Zhang, Li, Shen, Cai, Zhang, Chen, Dai, Ji, and Yang]{zhang2025vlaarena}
Borong Zhang, Jiahao Li, Jiachen Shen, Yishuai Cai, Yuhao Zhang, Yuanpei Chen, Juntao Dai, Jiaming Ji, and Yaodong Yang.
\newblock Vla-arena: An open-source framework for benchmarking vision-language-action models.
\newblock \emph{arXiv preprint arXiv:2512.22539}, 2025{\natexlab{a}}.

\bibitem[Zhang et~al.(2025{\natexlab{b}})Zhang, Hu, You, and Gao]{pmlr-v270-zhang25h}
Tong Zhang, Yingdong Hu, Jiacheng You, and Yang Gao.
\newblock Leveraging locality to boost sample efficiency in robotic manipulation.
\newblock In \emph{Proceedings of The 8th Conference on Robot Learning}, pages 3264--3284, 2025{\natexlab{b}}.

\bibitem[Zhang et~al.(2025{\natexlab{c}})Zhang, Qi, and Zheng]{zhang2025experiencesbenchmarkingvisionlanguageactionmodels}
Yihao Zhang, Yuankai Qi, and Xi~Zheng.
\newblock Experiences from benchmarking vision-language-action models for robotic manipulation.
\newblock \emph{arXiv preprint arXiv:2511.11298}, 2025{\natexlab{c}}.

\bibitem[Zhao et~al.(2025)Zhao, Qi, Hu, Zhu, Chen, Tian, Zhu, Howell, Huang, Walters, et~al.]{zhao2025generalizable}
Haibo Zhao, Yu~Qi, Boce Hu, Yizhe Zhu, Ziyan Chen, Heng Tian, Xupeng Zhu, Owen Howell, Haojie Huang, Robin Walters, et~al.
\newblock Generalizable hierarchical skill learning via object-centric representation.
\newblock \emph{arXiv preprint at arXiv:2510.21121}, 2025.

\bibitem[Zhou et~al.(2022)Zhou, Kumar, Finn, and Rajeswaran]{zhou2022policycompose}
Allan Zhou, Vikash Kumar, Chelsea Finn, and Aravind Rajeswaran.
\newblock Policy architectures for compositional generalization in control.
\newblock \emph{arXiv preprint arXiv:2203.05960}, 2022.

\bibitem[Zhou et~al.(2025)Zhou, Huang, Ni, Zhou, Yan, Yin, and Zhuo]{zhou2025factorhd}
Yifei Zhou, Xuchu Huang, Chenyu Ni, Min Zhou, Zheyu Yan, Xunzhao Yin, and Cheng Zhuo.
\newblock Factorhd: A hyperdimensional computing model for multi-object multi-class representation and factorization.
\newblock \emph{arXiv preprint at arXiv:2507.12366}, 2025.

\end{thebibliography}

\end{document}